\documentclass[sigconf]{acmart}
\usepackage{xcolor}
\usepackage{tcolorbox}
\usepackage{arydshln}
\usepackage{utfsym}
\usepackage{pifont}
\newcommand{\cmark}{\ding{51}}
\newcommand{\xmark}{\ding{55}}

\AtBeginDocument{%
  \providecommand\BibTeX{{%
    \normalfont B\kern-0.5em{\scshape i\kern-0.25em b}\kern-0.8em\TeX}}}

\newcommand{\toolname}{RAGProbe }    
\newcommand{\sfq}{A question combining multiple questions for which answers are in a single document}

\newcommand{\mfq}{A question combining multiple questions for which answers are in a set of documents}

\newcommand{\mc}{A question for which answer is not in the document corpus}

\newcommand{\mcq}{A multiple-choice question for which answer is in a single document}

\newcommand{\numberq}{A question to retrieve number for which answer is in a single document}

\newcommand{\datetime}{A question to retrieve date/time for which answer is in a single document}

\AtBeginDocument{%
  \providecommand\BibTeX{{%
    \normalfont B\kern-0.5em{\scshape i\kern-0.25em b}\kern-0.8em\TeX}}}

\setcopyright{acmlicensed}
\copyrightyear{2024}
\acmYear{2024}
\acmDOI{XXXXXXX.XXXXXXX}

\acmConference[Conference acronym 'XX]{Conference acronym 'XX}{June 03--05, 2024}{Woodstock, NY}
\begin{document}

%%
%% The "title" command has an optional parameter,
%% allowing the author to define a "short title" to be used in page headers.
\title{RAGProbe: An Automated Approach for Evaluating RAG Applications}

%RAGProbe: An Automated Approach for Evaluating RAG Applications
%RAGProbe: An Automated Approach for Testing RAG Applications
%RAGProbe: A Novel Approach for Testing RAG Applications
%RAGProbe: A Novel Approach for Testing Semantic Applications
%RAGProbe: A Novel Approach for Test Data Generation

\author{Shangeetha Sivasothy, Scott Barnett, Stefanus Kurniawan, Zafaryab Rasool, Rajesh Vasa}
\affiliation{%
\institution{Applied Artificial Intelligence Institute, Deakin University}
\city{Geelong}
\country{Australia}
}
\email{{s.sivasothy, scott.barnett, stefanus.kurniawan, zafaryab.rasool, rajesh.vasa}@deakin.edu.au}

%\renewcommand{\shortauthors}{Sivasothy et al.}

%%
%% By default, the full list of authors will be used in the page
%% headers. Often, this list is too long, and will overlap
%% other information printed in the page headers. This command allows
%% the author to define a more concise list
%% of authors' names for this purpose.
%\renewcommand{\shortauthors}{Trovato and Tobin, et al.}

%%
%% The abstract is a short summary of the work to be presented in the
%% article.
\begin{abstract}
Retrieval Augmented Generation (RAG) is increasingly being used when building Generative AI applications. Evaluating these applications and RAG pipelines is mostly done manually, via a trial and error process. Automating evaluation of RAG pipelines requires overcoming challenges such as context misunderstanding, wrong format, incorrect specificity, and missing content. Prior works therefore focused on improving evaluation metrics as well as enhancing components within the pipeline using available question and answer datasets. However, they have not focused on 1) providing a schema for capturing different types of question-answer pairs or 2) creating a set of templates for generating question-answer pairs that can support automation of RAG pipeline evaluation. In this paper, we present a technique for generating variations in question-answer pairs to trigger failures in RAG pipelines. We validate 5 open-source RAG pipelines using 3 datasets. Our approach revealed the highest failure rates when prompts combine multiple questions: 91\% for questions when spanning multiple documents and 78\% for questions from a single document; indicating a need for developers to prioritise handling these combined questions. 60\% failure rate was observed in academic domain dataset and 53\% and 62\% failure rates were observed in open-domain datasets. Our automated approach outperforms the existing state-of-the-art methods, by increasing the failure rate by 51\% on average per dataset. Our work presents an automated approach for continuously monitoring the health of RAG pipelines, which can be integrated into existing CI/CD pipelines, allowing for improved quality. 
\end{abstract}


%\begin{CCSXML}
%<ccs2012>
% <concept>
% <concept_id>00000000.0000000.0000000</concept_id>
%  <concept_desc>Do Not Use This Code, Generate the Correct Terms for Your Paper</concept_desc>
%<concept_significance>500</concept_significance>
% </concept>
% <concept>
%<concept_id>00000000.00000000.00000000</concept_id>
%<concept_desc>Do Not Use This Code, Generate the Correct Terms for Your Paper</concept_desc>
%<concept_significance>300</concept_significance>
%</concept>
% <concept>
%<concept_id>00000000.00000000.00000000</concept_id>
%  <concept_desc>Do Not Use This Code, Generate the Correct Terms for Your Paper</concept_desc>
%<concept_significance>100</concept_significance>
% </concept>
%<concept>
%<concept_id>00000000.00000000.00000000</concept_id>
% <concept_desc>Do Not Use This Code, Generate the Correct Terms for Your Paper</concept_desc>
%<concept_significance>100</concept_significance>
% </concept>
%</ccs2012>
%\end{CCSXML}

%\ccsdesc[500]{Do Not Use This Code~Generate the Correct Terms for Your Paper}
%\ccsdesc[300]{Do Not Use This Code~Generate the Correct Terms for Your Paper}
%\ccsdesc{Do Not Use This Code~Generate the Correct Terms for Your Paper}
%\ccsdesc[100]{Do Not Use This Code~Generate the Correct Terms for Your Paper}

%%
%% Keywords. The author(s) should pick words that accurately describe
%% the work being presented. Separate the keywords with commas.
%\keywords{LLMs, RAG, Testing, Hard Negatives, Information Retrieval}
%Software Testing

%AI Engineering

%% A "teaser" image appears between the author and affiliation
%% information and the body of the document, and typically spans the
%% page.

%\received{20 February 2007}
%\received[revised]{12 March 2009}
%\received[accepted]{5 June 2009}

%%
%% This command processes the author and affiliation and title
%% information and builds the first part of the formatted document.
\maketitle

\section{Introduction}

Retrieval Augmented Generation (RAG) has recently gained popularity for use in question and answer systems, as they better understand query context and meaning~\cite{jeong2024adaptive, yan2024corrective, asai2023self, es2023ragas, saad2023ares}. However, RAG has specific limitations and failure points~\cite{barnett2024seven, es2023ragas, jeong2024adaptive, wu2024faithful} due to imprecise embedding models and uncertainties introduced when generating results with large language models. Thus, questions a human takes for granted, a RAG pipeline may struggle with i.e. answering multiple questions from either a single document or a set of documents (see \autoref{tab:manual_execution} for more examples). These limitations are uncovered and potentially addressed through an evaluation of the a) quality of answers, b) impact of domain, and c) architectural choices. Improving RAG pipelines currently is often a manual and iterative process, that can be better supported by automation of the evaluation process - similar to automated test case generation.

\begin{table*}
\centering
\begin{tabular}
{|p{.02\linewidth}|p{.55\linewidth}|p{.04\linewidth}|p{.065\linewidth}|p{.065\linewidth}|p{.05\linewidth}|p{.08\linewidth}|}
\hline
\textbf{ID} & \textbf{Evaluation Scenario} & \textbf{Quivr} & \textbf{Danswer} & \textbf{Ragflow} & \textbf{Verba} & \textbf{Rag-stack} \\
\hline
S1 & \numberq & Pass & Pass & Pass & Pass & Fail \\
S2 & \datetime & Pass & Pass & Pass & Pass & Pass \\
S3 & \mcq & Fail & Pass & Pass & Fail & Pass \\
S4 & \sfq & Fail & Pass & Fail & Fail & Fail \\
S5 & \mfq & Fail & Partial & Fail & Fail & Fail \\
S6 & \mc & Fail & Partial & Fail & Fail & Fail \\
\hline
\end{tabular}
\caption{Effectiveness of manually created evaluation scenarios against open-source Retrieval Augmented Generation (RAG) solutions. Pass - evaluation scenario succeeded, Fail - evaluation scenario produced a defect, and Partial - answer correct but references incorrect.}
\label{tab:manual_execution}
\end{table*}

%Gaps from the existing literature
Prior work on exposing issues and addressing them focused on selecting better evaluation metrics~\cite{es2023ragas, saad2023ares, liu2023gpteval, fadnis2024inspectorraget} or improving RAG components by running them against existing question and answer datasets~\cite{mao2020generation, feng2024retrieval, zhang2024raft, salemi2024evaluating, yan2024corrective, jeong2024adaptive, asai2023self}. A recent work, RAGAS~\cite{es2023ragas} provides one such end-to-end pipeline for evaluating RAG pipelines, spanning data generation and  evaluation. However, RAGAS does not provide 1) a schema for capturing different types of question-answer pairs, or 2) a technique to generate question-answer pairs as templates (inspired by traditional test scenario templates) for evaluating a RAG pipeline. Our work is complimentary to RAGAS by addressing these limitations. The research gap identified is that there is no automated process for creating domain specific questions and answers covering a diverse set of variations. In this study, we address this.

In this work, we present, \toolname for automating the evaluation of RAG pipelines. We introduce the core concept of \textit{evaluation scenarios} to represent how to evaluate a RAG pipeline. An evaluation scenario when executed produces a question-answer pair and is designed to represent different types of variation found in question-answer pairs. An evaluation scenario includes a) a document corpus sampling and chunking strategy, b) scenario specific prompts and prompting strategy (for generation), and c) a set of evaluation metrics. To the best of our knowledge, this is the first study to identify and investigate variations in question-answer pairs for evaluation of RAG pipelines.  

We carried out an evaluation of \toolname across 5 open-source RAG pipelines, 3 datasets (Qasper, Google NQ, and MS Marco) and 180 question-answer pairs. Our study involved a) evaluating the effectiveness of our evaluation scenarios, b) comparing our approach with state-of-the-art, and c) analysing the impact of the domain. We found that the evaluation scenarios related to multiple questions posed in a single hit (S4 \& S5) resulted in the most failure rates (91\% and 78\%). We compared \toolname to RAGAS\cite{es2023ragas} and found that our approach produced more valid question-answer pairs (our approach produced 90\%, 98\%, and 92\% whereas the state-of-the -art approach resulted in 87\%, 93\%, and 85\% for Qasper, Google NQ, and MS Marco datasets respectively), and produced more failures across the RAG pipelines. This shows that our approach produces data that is more effective at evaluating RAG pipelines. 

We summarise our contributions as follows:
\begin{itemize}    
    \item An evaluation scenario schema for describing the constructs for evaluating RAG pipelines. 
    \item A set of evaluation scenarios that expose limitations in RAG pipelines. 
    \item A novel approach, \toolname for synthesising domain specific instances of evaluation scenarios based on a corpus. 
    \item An evaluation of \toolname using 3 public datasets and 5 open-source RAG pipelines.
    \item A set of recommendations from the literature on how to mitigate each of the identified evaluation scenarios. 
\end{itemize}

\section{Motivating Example} \label{sec:motivating_example}

Imagine Jack, a developer, building a RAG pipeline for a financial institution for question answering. This RAG pipeline enables users to ask questions based on a given corpus (a set of documents). RAG pipelines are required because the corpus includes proprietary documents and large language models (LLM) cannot reliably answer questions using their learned knowledge alone. To build a RAG pipeline, Jack needs to 1) load, parse and chunk the given corpus of documents, 2) index the chunks, 3) store chunks in a vector database (embedding store), and 4) write a prompt which instructs the large language model on how to generate an answer based on the retrieved chunks. To build an initial version of this RAG pipeline, Jack searches for open-source RAG pipelines and evaluates the functionality of existing RAG pipelines by indexing documents from publicly available datasets. 

Jack asks different variations of questions (e.g. number, date/time, multiple-choice questions, multiple questions combined in a single question) from 5 open-source RAG pipelines: 1) Quivr\footnote{https://github.com/QuivrHQ/quivr}, 2) Danswer\footnote{https://github.com/danswer-ai/danswer}, 3) Ragflow\footnote{https://github.com/infiniflow/ragflow}, 4) Verba\footnote{https://github.com/weaviate/Verba}, and 5) Rag-stack\footnote{https://github.com/psychic-api/rag-stack}. \autoref{tab:manual_execution} shows the results of manual execution of each question. All RAG pipelines failed to provide accurate responses for all evaluation scenarios (except S2). Jack realises the critical need to evaluate and improve the robustness of these pipelines to ensure they can handle a variety of evaluation scenarios successfully. Jack was wondering how to automatically generate question-answer pairs related to his domain, as the number of questions that can be asked from a given corpus is infinite and there is no systematic way to evaluate the developed RAG pipeline. Also, the manual approach is time-consuming. Therefore, Jack started looking at existing tools/approaches to generate question-answer pairs from documents and to consider variations of questions. \autoref{tab:comparison_of_tools} compares existing tools. Also, tracing failures within the RAG pipeline is challenging for Jack, particularly when determining which component — retrieval, indexer, prompt, LLM or generator — contributed to the failure. Lack of clear visibility into the system's internal workings complicates the process of identifying and resolving failures effectively.

%All RAG pipelines failed to answer 
%The fact that 5 out of 6 evaluation scenarios (excluding S2) failed in all the RAG pipelines indicates a high overall failure rate.
%%Jack realised that 2 out of the 6 evaluation scenarios did not give correct responses across all 5 open-source RAG pipelines. 

\begin{table*}
\centering
\begin{tabular}{|p{.32\linewidth}|p{.08\linewidth}|p{.08\linewidth}|p{.05\linewidth}|p{.09\linewidth}|p{.07\linewidth}|p{.06\linewidth}|p{.08\linewidth}|}
\hline
\textbf{Features} & \textbf{Corrective-RAG \cite{yan2024corrective}} & \textbf{Adaptive-RAG \cite{jeong2024adaptive}} & \textbf{ARES \cite{saad2023ares}} & \textbf{Inspector\newline RAGet \cite{fadnis2024inspectorraget}} & \textbf{Self-RAG \cite{asai2023self}} & \textbf{RAGAS \cite{es2023ragas}} & \textbf{Our \newline approach} \\
\hline
Generate question-answer pairs from documents & \xmark & \xmark & \cmark & \xmark & \xmark & \cmark & \cmark \\
Do not require an initial set of question-answer pairs & \cmark & \cmark & \xmark & \cmark & \cmark & \cmark & \cmark \\
Generate \& evaluate question-answer pairs automatically & \xmark & \xmark & \cmark & \cmark & \xmark & \cmark & \cmark \\
Show variations of questions & \xmark & \cmark & \xmark & \xmark & \cmark & \cmark & \cmark \\
Include schema for evaluation scenarios & \xmark & \xmark & \xmark & \xmark & \xmark & \xmark & \cmark \\
Use scenarios for question-answer generation & \xmark & \xmark & \xmark & \xmark & \xmark & \xmark & \cmark \\
\hline
\end{tabular}
\caption{Comparison of existing tools for evaluating RAG pipelines}
\label{tab:comparison_of_tools}
\end{table*}

%include schema for semantic tests

The motivating example highlights the following features that Jack requires for evaluating his application:
\begin{itemize}
    \item Generate domain and context specific questions and answers from a corpus for evaluating a RAG pipeline. 
    \item A schema for describing an evaluation scenario for a RAG pipeline. 
    \item A set of evaluation scenarios that cover a variety of question and answers used in a RAG pipeline.
    \item Automated evaluation of the answers produced by a RAG pipeline. 
\end{itemize}

% \item Techniques and strategies required for addressing or minimising identified limitations. 

\section{Evaluation Scenarios} \label{sec:test_scenarios}

In this paper, we define an evaluation scenario distinct from unit tests and test scenarios for the following reasons: a) unit tests and test scenarios, assume either a pass or failure outcome, b) neither cover the variance and nuance of natural language, and c) neither require custom evaluation metrics. The necessity for evaluation scenarios emerged from the process of implementing multiple RAG use cases, assessing open-source RAG implementations, and reviewing the relevant literature~\cite{barnett2024seven, rasool2024llms}.

\subsection{A schema for Evaluation Scenarios}

An evaluation scenario schema consists of 6 constructs: 1) document sampling strategy, 2) chunking strategy, 3) chunk sampling strategy, 4) scenario specific prompts, 5) a prompting strategy, and 6) acceptable evaluation metrics. 

\begin{itemize}
    \item \textit{Document sampling strategy:} Document sampling strategy indicates how documents need to be sampled from the corpus. These strategies can be random or iterative. Random sampling involves selecting documents from the corpus without a specific pattern. Iterative sampling means systematically going through every document in the corpus. 
    \item \textit{Chunking strategy:} This indicates how chunks are created from documents by specifying a particular strategy, separator, chunk size, and chunk overlap size. 
    \item \textit{Chunk sampling strategy:} This indicates how chunks are selected to generate questions. These strategies can be random or iterative. Random sampling involves selecting chunks from a document without a specific pattern. Iterative sampling means systematically going through every chunk of a document in the corpus.
    \item \textit{Scenario specific prompts:} Each evaluation scenario includes a specific prompt to generate question-answer pairs using LLMs. These prompts outline guidelines on dos and don'ts and specify the output format, such as a JSON object containing the question-answer pair.
    \item \textit{Prompting strategy:} Prompting strategy is the technique on how the prompt is written. These strategies can be one-shot (i.e. by specifying one example), three-shot (i.e. by specifying three examples), or sequential (i.e. by executing one prompt first and then execute the next prompt by passing the results of the first prompt into the context). 
    \item \textit{Acceptable evaluation metrics:} Different evaluation metrics are used to compare RAG generated answers against the expected response \cite{balaguer2024rag, es2023ragas, saad2023ares, fadnis2024inspectorraget, liu2023gpteval}. These metrics are: 1) correctness, 2) relevance, 3) completeness, 4) consistency, 5) explicitness, 6) contradiction, and 7) no-question related information. \textit{Correctness --} indicates the RAG generated response should match the expected answer exactly or be a close paraphrase, reflecting the same information. \textit{Relevance --} indicates the RAG generated response should directly address the question without deviating into unrelated topics. \textit{Completeness --} is where the RAG generated response covers all parts of the expected response, providing a full response. \textit{Consistency --} measures whether the RAG generated response aligns with the expected response in terms of detail and context, maintaining logical coherence. \textit{Explicitness --} indicates the RAG generated response should clearly state that the system does not know or cannot answer the question. \textit{Contradiction --} shows that the RAG generated response should not provide information or attempt to answer the question after stating the system cannot answer the question. \textit{No-question related information --} means that the RAG generated response should not include any information that attempts to address the question.
\end{itemize}

% Based on the experience from the AI Tutor \cite{barnett2024seven}, we identify 19 different scenarios as follows: 1) happy path (answer is directly available in the document), 2) happy path with answer being a list of items, 3) multiple questions based on a single document combined into a single question, 4) multiple questions based on multiple documents combined into a single question, 5) questions for which answer is not in the corpus, 6) questions which use abbreviations, 7) language variations (e.g. different slangs, colloquialism), 8) questions in a separate language other than the language of document corpus, 9) questions to retrieve numbers, 10) questions to retrieve numbers associated with International System of Units (e.g. miles, litres), 11) questions to retrieve location of an answer (e.g. page number of a document), 12) questions to retrieve date/time, 13) questions around summarizations, 14) questions around tables, 15) questions which have answers of yes or no, 16) questions to retrieve factual information, 17) multiple-choice questions, 18) jail breaking for developer mode questions, and 19) jail breaking for illegal activities. Jail breaking questions include developer mode questions, illegal activities, harmful content, fraudulent or deceptive activities, adult content, and unlawful practices. 

%Out of these 19 scenarios, 

%https://arxiv.org/pdf/2309.01431 - Benchmarking Large Language Models in Retrieval-Augmented Generation - 4 categories in this paper to add to a future data generation paper

\subsection{A set of Evaluation Scenarios}
We identify 6 illustrative scenarios to demonstrate the end-to-end automation of our approach. Three scenarios were selected based on the literature: 1) questions to retrieve numbers \cite{rasool2023evaluating}, 2) questions to retrieve date/time \cite{wu2024faithful}, and 3) multiple-choice questions \cite{rasool2023evaluating}. The remaining three scenarios were identified from a recent work that highlighted 7 failure points of RAG systems \cite{barnett2024seven}. These are: 4) combining multiple questions for which answers are in a single document, 5) combining multiple questions for which answers are in a set of documents, and 6) asking a question for which answer is not in the corpus. Examples of each scenario are shown in \autoref{tab:test_scenarios_examples} and we describe each scenario below. 

\begin{table*}
\centering
\begin{tabular}{|c|p{0.45\linewidth}|p{0.45\linewidth}|}
\hline
\textbf{ID} & \textbf{Question} & \textbf{Answer} \\
\hline
S1 & What was the total attendance for the 1982 World Series? & 384,570 \\
S2 & When was the National Insurance Scheme introduced in Jamaica? & 1966 \\
S3 & Which nation hosted the 2014 Winter Paralympics? A): United States B): Canada C): Russia D): Germany & C \\
S4 & What does Charles Hockett identify as a core feature of human language? What is the method used to quantify the property of a communication system to combine and re-use elementary forms in a lexicon? What coding was applied to subdivide each word of the lexicon in the phonetic analysis? & Charles Hockett identified duality of patterning as a core feature of human language. A measure called `combinatoriality', which is a real-valued quantity ranging in the interval [0 : 1] that quantifies how frequently forms recur in a lexicon, is used to quantify this property. The International Phonetic Association (IPA) coding was applied to subdivide each word of the lexicon in the phonetic analysis. \\
S5 & Where did the 1924 Winter Olympics take place? Who produced track 3 of the album? Who was the first member of the 50 home run club? & The 1924 Winter Olympics took place in Chamonix. Track 3 of the album was produced by Vinylz, Boi-1da and Velous. Babe Ruth was the first member of the 50 home run club. \\
S6 & Explain quantum mechanics? & Sorry, the system doesn't know the answer. \\
\hline
\end{tabular}
\caption{Overview of evaluation scenarios with example question-answer pairs}
\label{tab:test_scenarios_examples}
% \vspace{-5mm}
\end{table*}

\textbf{S1: \numberq}
%\textbf{S5: Asking a question to retrieve number} 

In this scenario, a question is formulated with the expectation of receiving a numerical value as a response. The RAG pipeline is expected to extract relevant numerical information from a document to provide an accurate numeric response. The evaluation scenario schema is as follows:
\begin{itemize}
    \item Document sampling strategy - Random N documents (e.g. N=10)
    \item Chunking strategy - Character text splitting, separator=newline, chunk size=3000, chunk overlapping size=150
    \item Chunk sampling strategy - Random one chunk per document after filtering chunks containing numbers
    \item Scenario specific prompts - Specific prompt
    \item Prompting strategy - One-shot
    \item Acceptable evaluation metrics - Correctness, relevance, completeness, consistency 
\end{itemize}

%For example, when considering a document\footnote{\url{https://en.wikipedia.org/wiki/1982_Milwaukee_Brewers_season}}, a question is asked for which the answer has a numerical value. An example question is \textit{What was the total attendance for the 1982 World Series?}. 

%How many conditions can the MRI machine be used for?
%\url{https://www.nhlbi.nih.gov/health/heart-tests}

%Langchain character text splitting\footnote{\url{https://python.langchain.com/v0.1/docs/modules/data_connection/document_transformers/character_text_splitter/}}

\textbf{S2: \datetime}
%\textbf{S6: Asking a question to retrieve date/time}

In this scenario, a question is asked to obtain temporal information, specifically a date or time, as the response. The system is expected to retrieve relevant date or time-related details from the document corpus to provide an accurate temporal response. The evaluation scenario schema is as follows:
\begin{itemize}
    \item Document sampling strategy - Random N documents (e.g. N=10)
    \item Chunking strategy - Character text splitting, separator=newline, chunk size=3000, chunk overlapping size=150
    \item Chunk sampling strategy - Random one chunk per document after filtering chunks containing date/time
    \item Scenario specific prompts - Specific prompt
    \item Prompting strategy - One-shot
    \item Acceptable evaluation metrics - Correctness, relevance, completeness, consistency 
\end{itemize}

%For example, when considering a document\footnote{\url{https://www.soa.org/library/newsletters/pension-section-news/2007/may/pen07may/}}, a question is asked for which the answer has date/time. An example question is \textit{When was the National Insurance Schema introduced in Jamaica?} 

%How many conditions can the MRI machine be used for?
%What was the total attendance for the 1982 World Series? - 384,570
%How many goals did the Canadian ice-hockey team score in total during their qualifying round? - 85

%How many goals did the Canadian ice-hockey team score in total during their qualifying round?

\textbf{S3: \mcq}
%\textbf{S4: Asking a multiple-choice question}

In this scenario, a question is created along with several predetermined answer options. The system is expected to select the correct answer option from the provided options based on its understanding of the question and the context provided in the document corpus. Multiple-choice questions often require the system to analyze and interpret information to discern the most appropriate response among the given choices. The evaluation scenario schema is as follows:
\begin{itemize}
    \item Document sampling strategy - Random N documents (e.g. N=10)
    \item Chunking strategy - Character text splitting, separator=newline, chunk size=3000, chunk overlapping size=150
    \item Chunk sampling strategy - Random one chunk per document
    \item Scenario specific prompts - Specific prompt
    \item Prompting strategy - Three-shot
    \item Acceptable evaluation metrics - Correctness, relevance, completeness, consistency 
\end{itemize}

%For example, when considering a document\footnote{\url{https://en.wikipedia.org/wiki/2014_Winter_Paralympics}}, a multiple-choice question is asked from the available content, by providing four options, one of which is the correct answer. An example question is \textit{Which nation hosted the 2014 Winter Paralympics? A): United States B): Canada C): Russia D): Germany}.

\textbf{S4: \sfq}

This scenario involves formulating a single question by combining multiple individual questions that all relate to information contained within a single document. The combined question requires the system to synthesise answers from different parts of the same document to provide a response. The system response is considered to be complete if all questions inside a single question have been answered. The evaluation scenario schema is as follows:

%For example, when considering a document (e.g. research paper)\footnote{\url{https://arxiv.org/pdf/1602.03661}}, three distinct paragraphs are chosen from various sections, and a question is formulated based on each paragraph. Subsequently, these three questions are combined and presented as a single question. The example question is \textit{What does Charles Hockett identify as a core feature of human language? What is the method used to quantify the property of a communication system to combine and re-use elementary forms in lexicon? What coding was applied to subdivide each word of the lexicon in the phonetic analysis?}  

\begin{itemize}
    \item Document sampling strategy - Random N documents (e.g. N=10)
    \item Chunking strategy - Character text splitting, separator=newline, chunk size=3000, chunk overlapping size=150
    \item Chunk sampling strategy - Random three chunks from a single document combined
    \item Scenario specific prompts - Specific prompt
    \item Prompting strategy - One-shot
    \item Acceptable evaluation metrics - Correctness, relevance, completeness, consistency 
\end{itemize}

\begin{figure*}
\centering
\includegraphics[width=0.8\linewidth]{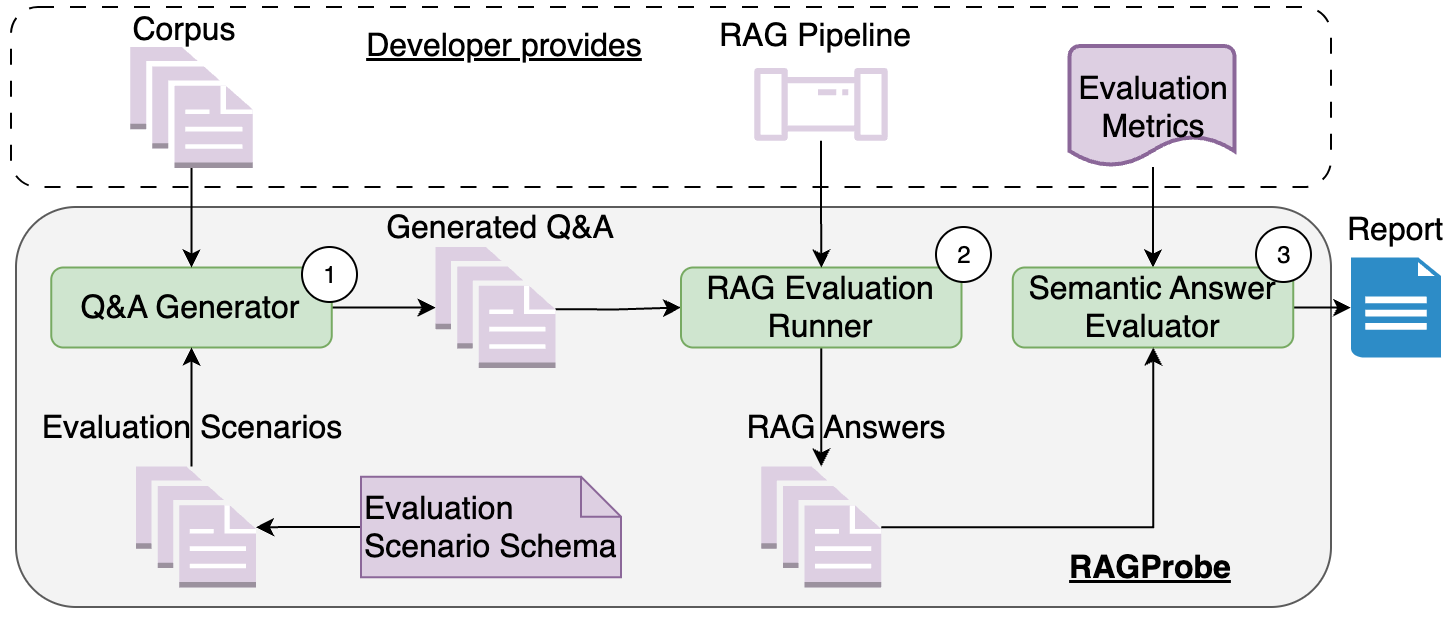}
\caption{RAGProbe: Our automated approach to generate question-answer pairs. Our approach is extensible by adding different evaluation scenarios and different evaluation metrics.}
\label{fig:proposed_approach}
\end{figure*}

\textbf{S5: \mfq}
%\textbf{S2: Asking a question by combining multiple questions from multiple documents}

This scenario involves creating a single question by combining multiple individual questions, each sourced from different documents. The consolidated question requires the system to combine information from various sources to provide a response. The system response is considered to be complete if all questions inside a single question have been answered. The evaluation scenario schema is as follows:
\begin{itemize}
    \item Document sampling strategy - Random N documents (e.g. N=10)
    \item Chunking strategy - Character text splitting, separator=newline, chunk size=3000, chunk overlapping size=150
    \item Chunk sampling strategy - Random three chunks from a set of documents combined
    \item Scenario specific prompts - Specific prompt
    \item Prompting strategy - One-shot
    \item Acceptable evaluation metrics - Correctness, relevance, completeness, consistency 
\end{itemize}

%As an example, when considering three documents\footnote{\url{https://en.wikipedia.org/wiki/1924_Winter_Olympics}}, \footnote{\url{https://en.wikipedia.org/wiki/4_Your_Eyez_Only}}, \footnote{\url{https://en.wikipedia.org/wiki/50_home_run_club}}, a single paragraph is selected from each document, and a question is formulated based on each chosen paragraph. Subsequently, these three questions are combined and presented as a single question. The example question is \textit{Where did the 1924 Winter Olympics take place? Who produced track 3 of the album? Who was the first member of the 50 home run club?}.

\textbf{S6: \mc}
%\textbf{S3: Asking a question for which the corpus does not have an answer}

In this scenario, a question is posed to the system for which there is no corresponding answer available within the corpus of documents that the system has access to. Therefore, the system is expected to respond that it doesn't know the response. The evaluation scenario schema is as follows:
\begin{itemize}
    \item Document sampling strategy - Random N documents (e.g. N=10)
    \item Chunking strategy - Character text splitting, separator=newline, chunk size=3000, chunk overlapping size=150
    \item Chunk sampling strategy - Iterative throughout the entire corpus
    \item Scenario specific prompts - Specific prompt
    \item Prompting strategy - Sequential (i.e. one prompt executed after the other prompt)
    \item Acceptable evaluation metrics - Explicitness, contradiction, no-question related information 
\end{itemize}

%For example, when a document corpus has documents\footnote{\url{https://en.wikipedia.org/wiki/1924_Winter_Olympics}}, \footnote{\url{https://en.wikipedia.org/wiki/4_Your_Eyez_Only}}, \footnote{\url{https://en.wikipedia.org/wiki/50_home_run_club}}, a question \textit{Explain quantum mechanics?} is asked where the answer is not available in the document corpus.

\section{\toolname}

In this section, we describe our proposed approach, RAGProbe along with a usage example.

\subsection{Our Approach}

To address the issues discussed in the motivating example (\autoref{sec:motivating_example}), we propose a novel approach, RAGProbe, that leverages LLMs for generating context specific evaluation scenarios and evaluates a given RAG pipeline. As shown in \autoref{fig:proposed_approach}, our approach has three key components: 1) Q\&A Generator, 2) RAG Evaluation Runner, and 3) Semantic Answer Evaluator. The Q\&A Generator takes an existing document corpus as an input, and then applies evaluation scenario schemas to generate questions and answers. The RAG Evaluation Runner is responsible for adapting to the RAG implementation (handle authentication and mapping to API) and collecting answers to all generated questions from the RAG pipeline under evaluation. Semantic Answer Evaluator compares generated answers by Q\&A Generator against RAG pipeline generated answers. To handle the ambiguity in natural language responses, we extend OpenAI evals ClosedQA template\footnote{https://github.com/openai/evals/blob/main/evals/registry/modelgraded/closedqa.yaml} to determine whether the RAG generated response is correct based on the expected response. 

\toolname is implemented in Python and is available online\footnote{https://figshare.com/s/e0d74c0d346fd2e05d59}. We use GPT-4 \cite{openai2023gpt4} to generate question-answer pairs for scenarios S1-S5, enhancing the quality of the generated pairs. For scenario S6, we use GPT-3.5-Turbo to reduce costs, as S6 involves iterative sampling of chunks and repeated application of the LLM.

%These answers need to be compared with the generated answers. This task is performed by the Semantic Answer Evaluator. 

%We use GPT-4 \cite{openai2023gpt4} to generate question-answer pairs for the scenarios S1-S5 for enhancing the quality of generated question-answer pairs. We use GPT-3.5-Turbo for scenario S6 to reduce cost, as S6 uses iterative sampling of chunks and the iterative application of the LLM is expensive. 

%We use GPT-3.5-Turbo for scenario S6 to reduce cost (iterative application of the LLM is expensive). 

% , and also we send 8 consecutive chunks inside one prompt to further reduce cost. 

% RAG Evaluation Runner executes generated questions for the given RAG pipeline. Finally, Semantic Answer Evaluator compares the RAG generated responses against the expected responses for the given evaluation criteria and creates a report on succeeded and failed questions. 
%usage example needs to be rewritten - not clear
\subsection{Usage Example}
\toolname is designed to be used by developers in three ways: 1) to generate question-answer pairs, 2) to enable evaluation of a RAG pipeline, and 3) add new evaluation scenarios. For generating question-answer pairs, developers provide a corpus of documents (e.g. a set of PDFs). Then, \toolname generates a set of questions and answers. The end-to-end evaluation requires developers to provide access to a RAG pipeline (i.e. access to the database and search interface) and to a set of generated questions. Adding new scenarios requires filling out the evaluation schema: selecting sampling strategies, chunking strategy, and creating the prompts for use in a prompting strategy.

As shown in \autoref{fig:proposed_approach}, Q\&A Generator of \toolname will use the corpus of documents, provided by developers, to generate question-answer pairs. RAG Evaluation Runner takes the RAG pipeline, given by developers, adapts to its implementation (e.g. authentication, mapping to API) to generate answers for generated questions by Q\&A Generator. Semantic Answer Evaluator compares the answers generated by Q\&A Generator against the answers generated by the given RAG pipeline, and provides a report on passing and failing questions. Semantic Answer Evaluator allows developers to substitute other evaluation metrics. 

%For generating question-answer pairs, (1) developers provide a corpus of documents (i.e. a set of PDFs) and (2) a set of questions and answers are generated. The end-to-end evaluation is to provide recommendations for how to mitigate or reduce the identified issues. (3) The end-to-end evaluation requires providing access to a RAG implementation (i.e. access to the database and search interface) and to a set of generated questions. Adding new scenarios requires filling out the evaluation schema: selecting sampling strategies, chunking strategy, and creating the prompts for use in a prompting strategy. 

By using RAGProbe (as shown in \autoref{fig:proposed_approach}), Jack is able to: 1) generate questions and answers relevant to his domain, 2) have a schema of evaluation scenarios, 3) have a set of evaluation scenarios to evaluate a RAG pipeline, and 4) have an automated way to evaluate RAG during development or as part of a CI/CD pipeline.

\section{Evaluation of \toolname}

%test sceanrios replaced with evaluation scenarios 
We evaluated \toolname to assess: 1) the effectiveness of evaluation scenarios, 2) how \toolname compares to state-of-the-art approaches, and 3) the impact of datasets from different domains. We present the following research questions for our evaluation:

\textbf{RQ1: How effective are the evaluation scenarios in exposing failure rates in open-source RAG pipelines?} \newline
Unlike state-of-the-art techniques that randomly generate question-answer pairs, \toolname generates questions for specific scenarios relevant to RAG pipelines. We first evaluate these scenarios by analysing how effective they are at exposing limitations in 5 open-source RAG pipelines. 

%question and answer systems

\textbf{RQ2: How does our approach compare to existing state-of-the-art approaches in terms of failure rate and validity?} \newline
We compare \toolname with the state-of-the-art approach (RAGAS \cite{es2023ragas}) found in the literature in terms of how effective the approaches are for finding defects (failure rate) and the quality of the generated questions in terms of validity to the scenario. We did not find other tools that provide end-to-end evaluation of RAG and question and answer generation from a corpus. 

%RQ3: What is the impact of different domains on the effectiveness of automatically generated test data?
\textbf{RQ3: What is the impact of different domains on the effectiveness of automatically generated question-answer pairs?} \newline
Our hypothesis is that different domains use different terms which will impact the quality of the evaluation scenarios (i.e. domains will make it harder to find defects). This question examines the influence of domain-specific concepts (found in different datasets) for the different scenarios.

%Can we use the term test data
\subsection{Datasets}
For our evaluation, we needed two different datasets, a) a set of corpus of documents, and b) a set of RAG implementations. 

\subsubsection{Corpus Datasets}

%Brief overview of datasets
For a set of corpus, we use i) Qasper \cite{dasigi2021dataset}, ii) Google Natural Questions (NQ) \cite{kwiatkowski2019natural}, and iii) MS Marco \cite{nguyen2016ms}. Qasper is a dataset of 1,585 scientific research papers. Google NQ corpus is a question answering dataset with 307,373 training examples. These documents are from Wikipedia pages. MS Marco is a dataset of real web documents using the Bing search engine. MS Marco dataset contains 3.2 million documents and 8.8 million passages. %file format
From all three datasets, we used documents that could be downloaded and indexed by our selected RAG pipelines (see \autoref{sec:rag}). We noticed that some URLs in MS Marco dataset were obsolete and no longer valid, so we excluded those URLs when downloading documents. Qasper dataset contains of PDF files, Google NQ dataset contains of HTML files, and MS Marco dataset contains a combination of PDF and HTML files. %domains of datasets
With respect to domains, Qasper dataset belongs to academic or scientific research domain which has scientific research papers. Both Google NQ and MS Marco datasets are open-domain, meaning they do not pertain to any specific domain. All documents were converted to plain text before ingestion. 

%\autoref{tab:summary_datasets} provides a summary of these datasets. 
%Natural Language Processing
%Google NQ includes 3.2 million documents, each document being a Wikipedia page. - Google NQ  with single annotations, 7,830 examples with 5-way annotations for development data, and a further 7,842 examples 5-way annotated as test data

%\begin{table}[htb]
%\centering
%\begin{tabular}{|p{.18\linewidth}|p{.25\linewidth}|p{.2\linewidth}|p{.18\linewidth}|}
%\hline
%\textbf{Dataset} & \textbf{Domain} & \textbf{Doc Types} & \textbf{Doc Count} \\
%\hline
%Qasper & Academic & PDF & 1,585 \\
%Google NQ & Open-domain & HTML & 307,373 \\
%MS Marco & Open-domain & PDF, HTML & 3,200,000 \\
%\hline
%\end{tabular}
%\caption{Summary of datasets used in the study}
%\label{tab:summary_datasets}
%\end{table}

%\textbf{Description}
%Description
%Dataset - 
%Qasper - Scientific papers with questions
%MS Marco - Web search queries from Bing and related URLs
%Google NQ - Web search queries from Google and related wikipedia pages

\subsubsection{Open-source RAG Repositories}
\label{sec:rag}
We selected 5 open-source RAG repositories from Github (see \autoref{tab:rag_repositories}). Our selection criteria for RAG repositories was 1) contains keywords `retrieval augmented generation' or `rag', 2) timeline (the repository must be created in the last 5 years), 3) popularity (the repository must have at least 1000 stars), 4) activeness (the repository must have at least a commit in the last 2 years), 5) programming language (the repository doesn’t necessarily have to be Python-specific), 6) functionality (the repository should support uploading documents,  the repository should not test RAG pipelines, and the repository does not require code changes to upload documents), 7) purpose (the repository should not be libraries or library wrappers or API wrappers), and 8) language (the repository description should be written in English). 

\begin{table*}
\centering
\begin{tabular}
{|p{.1\linewidth}|p{.05\linewidth}|p{.2\linewidth}|p{.25\linewidth}|p{.20\linewidth}|}
\hline
\textbf{Repo Name} & \textbf{Stars} & \textbf{Programming Language} & \textbf{Large Language Model} & \textbf{Embedding} \\
\hline
Quivr & 31,800 & Python & gpt-3.5-turbo & text-embedding-ada-002  \\
Danswer & 9,000 & Python & gpt-3.5-turbo-16k-0613 & intfloat/e5-base-v2\\
Ragflow & 4,400 & Python & gpt-3.5-turbo & text-embedding-ada-002 \\
Verba & 2,100 & Python & gpt-4-1106-preview & text-embedding-ada-002 \\
Rag-stack & 1,400 & Python & ggml-gpt4All-J-v1.3-groovy & all-MiniLM-L6-v2\\
\hline
\end{tabular}
\caption{Overview of the open-source RAG pipelines with default settings. Note: our approach is a black box testing approach.}
\label{tab:rag_repositories}
% \vspace{-5mm}
\end{table*}

The search term ``retrieval augmented generation" yielded 1.2k projects, while the term ``RAG" returned 36.4k projects. Out of these GitHub projects, repositories with more than 1000 stars and created after 2019, resulted in 53 repositories. Out of these 53 repositories, project description was written in English for 50 repositories. We manually inspected each of these repositories to check whether 1) the repository supports uploading documents, 2) the repository does not require changing code to upload documents, 3) the repository does not test RAG pipelines, and 4) the repository is not a library or a library wrapper or an API wrapper (For example, Haystack\footnote{https://github.com/deepset-ai/haystack}, RAGatouille\footnote{https://github.com/bclavie/RAGatouille} were excluded because they were libraries). Our selection criteria resulted in 6 repositories, and we manually set up all these 6 repositories for evaluation. We failed to set up one repository, superagent\footnote{https://github.com/superagent-ai/superagent} due to complexity with indexing documents. This resulted in 5 RAG repositories, which are 1) Quivr, 2) Danswer, 3) Ragflow, 4) Verba, and 5) Rag-stack. \autoref{tab:rag_repositories} shows the summary of the selected RAG repositories for evaluation. We used default settings of these repositories to execute the generated questions. 

%because it required calling APIs to index documents and was not properly documented

%Quivr\footnote{https://github.com/QuivrHQ/quivr}, 2) Danswer\footnote{https://github.com/danswer-ai/danswer}, 3) Ragflow\footnote{https://github.com/infiniflow/ragflow}, 4) Verba\footnote{https://github.com/weaviate/Verba}, and 5) Rag-stack\footnote{https://github.com/psychic-api/rag-stack}. 

%Ragflow - %Qwen-plus 
%Framework
%Quivr - Langchain, Llama Index
%Danswer - Langchain, Llama Index
%Ragflow - 
%Verba - 
%Ragstack - Langchain

%\subsection{Semantic Test Data Synthesis}

\subsection{Methodology}

To answer RQ1, we random sampled 10 documents from each dataset of Qasper, Google NQ, and MS Marco for generation of question-answer pairs. Then, we converted the selected documents into a text file (both PDF and HTML files). We automatically generated 30 questions for each evaluation scenario. This resulted in 180 questions to be executed across one RAG pipeline. Then, we captured the generated responses and evaluated against the expected responses as per the evaluation metrics. Finally, we calculated the failure rate per evaluation scenario. Further, we break down failure rate per evaluation scenario per RAG pipeline. \autoref{tab:failure_question} shows an example of failure with a question and answer.

%of 1) a question to retrieve number for which answer is in a single document, 2) a question to retrieve date/time for which answer is in a single document, 3) a multiple-choice question for which answer is in a single document, 4) a question combining multiple questions for which answers are in a single document, 5) a question combining multiple questions for which answers are in a set of documents, and 6) a question for which answer is not in the document corpus. 

%We define a failure, in the context of RAG, as any instance where the RAG pipeline does not successfully retrieve and generate accurate, relevant, or contextually appropriate responses to a given query for the given scenarios. It is determined based on the ground-truth (expected responses) using a set of evaluation metrics.}

%Note that since our approach aims to identify failures in a RAG pipeline, a higher failure rate is more desirable. 

%Please refer to a library database, a book retailer, or the publisher's information for the ISBN number of the book.

%An example of failure is show
\begin{table}
\centering
\begin{tabular}
{|p{.23\linewidth}|p{.7\linewidth}|}
\hline
% & \textbf{Example} \\
%\hline
Question & What is the ISBN number of the book `The Unity of Science and Economics: A New Foundation of Economic Theory'? \\
\hdashline
Expected Response & 978-1-4939-3464-5 \\
\hdashline
RAG Response & The provided context does not contain any information about the ISBN number of the book ``The Unity of Science and Economics: A New Foundation of Economic Theory." Thus, I am unable to locate the specific ISBN you are looking for. Please refer to a library database, a book retailer, or the publisher's information for the ISBN number of the book. \\
\hdashline
Failed Evaluation Metric & Correctness \\
\hline
\end{tabular}
\caption{An example of failure with a question and answer }
\label{tab:failure_question}
\vspace{-5mm}
\end{table}
\begin{tcolorbox}
[width=3.3in, size=fbox, sharp corners, float]
{Failure is any answer from the RAG pipeline that does not pass one or more evaluation metrics. The LLM is used to assess pass/fail for each criteria given the ground-truth and the generated-answer.} \end{tcolorbox}

%Failure rate is the ratio of the number of failures of a given category to a given unit of measure\footnote{\url{https://istqb-glossary.page/}}. 

%\numberq, 2) \datetime, 3) \mcq, 4) \sfq, 5) \mfq, and 6) \mc.

%Note that as our approach is intended to find failures in a RAG pipeline, the higher the better. 

%In total, there were 180 test cases executed across each RAG pipeline, 150 test cases per test scenario, and 300 test cases per dataset.

% mentioned in \autoref{subsec:semantic_answer_evaluator}. 

%1) asking a question by combining multiple questions from a document, 2) asking a question by combining multiple questions from multiple documents, 3) asking a question for which the corpus does not have an answer, 4) asking a multiple-choice question, 5) asking a question to retrieve number, and 6) asking a question to retrieve time. 

To answer RQ2, we compared our approach to a state-of-the-art approach, RAGAS \cite{es2023ragas}. We initially searched the literature for tools used to evaluate RAGs, focusing specifically on those that generate questions from documents. Then, we generated questions using the selected tools and executed against the selected 5 open-source RAG pipelines. We compared failure rate against our approach versus the state-of-the-art approach. From the generated questions, we performed a statistical test (the Wilcoxon Signed-Rank Test \cite{woolson2005wilcoxon}) to determine if there is a significant difference between the failure rates between the two approaches. 

We evaluated the validity of the generated questions by comparing chunks, a method which had previously been used in the literature \cite{chen2024few}. We used similar chunking strategies from our approach and from the state-of-the art approaches to store in a vector store for the selected documents. Then, we used generated questions to retrieve relevant chunks using a retriever from the vector store. We excluded questions generated from S6 (a question for which answer is not in the document corpus) as they didn't have pre-assigned chunks. Then, we compared whether the retrieved chunks and the pre-assigned chunks for generated questions were similar. We then calculated the percentage of similarity between the retrieved chunks and the pre-assigned chunks for generated questions from both our approach and the state-of-the-art approach.

To answer RQ3, we group generated questions per dataset. There were 10 questions per dataset for a given scenario. In total, there were 300 questions (10 questions per dataset * 6 scenarios * 5 RAG pipelines). Then, we calculated the total number of failing questions per dataset. Further, we calculated the total number of RAG pipelines, that failed to answer all 10 questions from a dataset of a given scenario. All data and results are available online\footnote{https://figshare.com/s/e0d74c0d346fd2e05d59}.

%analysed results, obtained from RQ1, across three datasets.

%Follow up study: RAG pipelines have become a popular search method with many reference architectures and implementations available. However, the effectiveness of the different architectural choices has yet to be established and requires SEs to manuall evaluate each one. In this study we carry out a thorough empirical evaluation of RAG pipelines and find that A, B and C are the most effective implementation choices for a RAG

%Figshare DOI - 10.6084/m9.figshare.25940956
%\footnote{https://github.com/a2i2/rag-pipelines}. % - To be replaced with figshare link

\subsection{Results}

\subsubsection{How effective are the evaluation scenarios in exposing failure rates in open-source RAG pipelines?} 

\begin{figure}[h]
    \centering
    \includegraphics[width=\linewidth]{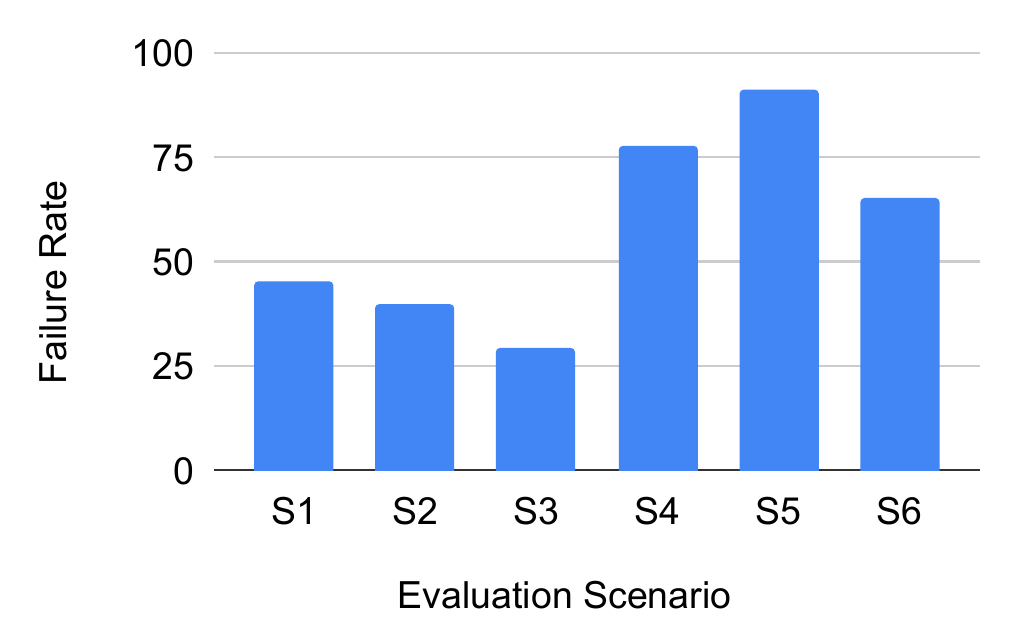}
    \caption{Total failure rate combining all 5 RAG pipelines. The failure rate is calculated as the number of failures divided by the total number of questions.}
    \label{fig:overall_evaluation}
    % \vspace{-5mm}
\end{figure}

\autoref{fig:overall_evaluation} shows the summary of automatic execution of automatically generated questions. In total, there were 150 generated questions per scenario, that were executed across 5 RAG pipelines. Scenario S5 has the highest failure rate at 91\%, followed by S4 at 78\% and S6 at 65\%, indicating these scenarios are the most problematic. Scenarios S1 and S2 show moderate failure rates at 45\% and 40\% respectively. Scenario S3 has the lowest failure rate at 29\%, suggesting it encounters the fewest number of failing questions across 5 RAG pipelines.

%Failed test cases per scenario
As shown in \autoref{tab:total_percentage_rag_pipeline_test_scenario}, Quivr and Rag-stack exhibited the highest failure rates (100\%) in scenarios S4 and S5. When asking questions for which the answer is not in the corpus, Danswer did not generate any correct responses, but performed well in multiple-choice questions. During the manual inspection, we found that Danswer generated responses by referring to the large language model's trained knowledge, instead of answering ``The system doesn't know the answer". Ragflow had the lowest failure rate (27\%) with multiple-choice questions compared to all other scenarios for this RAG pipeline. Verba had the lowest failure rate (17\%) when handling multiple-choice questions compared to all other scenarios for this RAG pipeline. Compared to 5 RAG pipelines, Rag-stack had the highest failure rates in multiple scenarios (S1, S2, S3, S4, S5), indicating significant challenges in handling complex questions and specific information retrieval.

%In total, there were 180 generated questions that were executed for each RAG pipeline. Overall, Quivr and Verba had the lowest failing questions (81/180) whereas Rag-stack had the highest number of failing questions (150/180). Two RAG pipelines (Quivr and Rag-stack) didn't generate any correct responses when multiple questions for which answers are in a single document and in a set of documents were asked as a single question. In total, there were 150 generated questions that were executed across 5 RAG pipelines, for a given scenario. Overall, 78\% (117/150) of questions failed to generated correct responses when multiple questions from a single document were asked as a single question. 91\% (137/150) of questions failed to generated correct responses when multiple questions from a set of documents were combined and asked. Overall, 65\% (98/150) of incorrect responses were generated when questions for which the answer is not in the corpus were asked to the RAG pipelines. 

%\begin{figure}[h]
%    \centering
%    \includegraphics[width=\linewidth]{Figures/Overall_Evaluation_Latest.png}
%    \caption{Automatic test case generation and execution across 5 RAG pipelines and 3 datasets, combining all 6 evaluation scenarios.}
%    \label{fig:overall_evaluation}
%    \vspace{-5mm}
%\end{figure}

%For the evaluation scenario of retrieving numbers, there were 55\% (82/150) of correct responses. There were 60\% (90/150) of correct responses for the evaluation scenario of retrieving date/time. When compared against all evaluation scenarios, the number of correct responses were the highest (71\%) for the evaluation scenario of asking multiple-choice questions. Overall, the lowest (9\%) number of correct responses were obtained for the evaluation scenario of when asking a question by combining multiple questions from multiple documents. Further, we observed that not all RAG pipelines provide references (i.e. from where answers are retrieved). 

%%As shown in \autoref{fig:evaluation_number}, 
% %, as shown in \autoref{fig:evaluation_date_time}. 
%%, as can be seen in \autoref{fig:evaluation_mcq}. 

\begin{table}[h]
\centering
\begin{tabular}   
{|p{.03\linewidth}|p{.1\linewidth}|p{.15\linewidth}|p{.15\linewidth}|p{.1\linewidth}|p{.20\linewidth}|}
\hline
\textbf{ID} & \textbf{Quivr} & \textbf{Danswer} & \textbf{Ragflow} & \textbf{Verba} & \textbf{Rag-stack}\\
\hline
S1 &  17\% & 43\% & 50\% & 43\% & 73\% \\
S2 &  20\% & 30\% & 43\% & 30\% & 77\% \\
S3 &  17\% & 17\% &	27\% &	17\% & 70\% \\
S4 &  100\% & 70\% & 80\% &	40\% & 100\% \\
S5 &  100\% & 70\% & 97\% & 90\% & 100\%  \\
S6 &  17\% & 100\% & 80\% & 50\% & 80\% \\
\hline
\end{tabular}
\caption{Failure rates of RAG pipelines across evaluation scenarios}
\label{tab:total_percentage_rag_pipeline_test_scenario}
\vspace{-5mm}
\end{table}

%However, Quivr had the least number of failed test cases for handling questions for which the answer is not in the corpus and for questions to retrieve numbers.

%Verba had the lowest failed test cases in handling combined questions from a single document. Ragflow had the highest failure rates in multiple scenarios (S1, S2, S3, S4, S5), indicating significant challenges in handling complex questions and specific information retrieval.

\begin{tcolorbox}
[width=3.3in, center upper, size=fbox, sharp corners, float]
{\textbf{Answer to RQ1:} All evaluation scenarios had a total failure rate between 29\% and 91\% for all 30 questions across 5 RAG pipelines. Asking multiple-choice questions exhibited the lowest failure rate, whilst asking questions by combining multiple questions from a set of documents had the highest.}\end{tcolorbox}

%%Overall, the RAG pipelines exhibit the lowest failure rates with multiple-choice questions and retrieving specific date/time information, while they struggle the most with complex questions that involve synthesising information from multiple sources.

\subsubsection{How does our approach compare to existing state-of-the-art approaches in terms of failure rate and validity?}

By reviewing the literature, we identified two tools capable of generating question-answer pairs from documents: RAGAS \cite{es2023ragas} and ARES \cite{saad2023ares}. When setting up these tools, ARES \cite{saad2023ares} required an initial set of question-answer pairs. Therefore, we used RAGAS \cite{es2023ragas} to generate question-answer pairs for comparison with our approach. Out of the 180 questions generated by RAGAS, 24 had answers listed as ``nan" (i.e., without valid answers). This indicates that RAGAS has limitations in accurately generating valid answers, affecting the utility of the generated data for downstream tasks. 

\begin{figure}[htb]
\centering
\includegraphics[width=\linewidth]{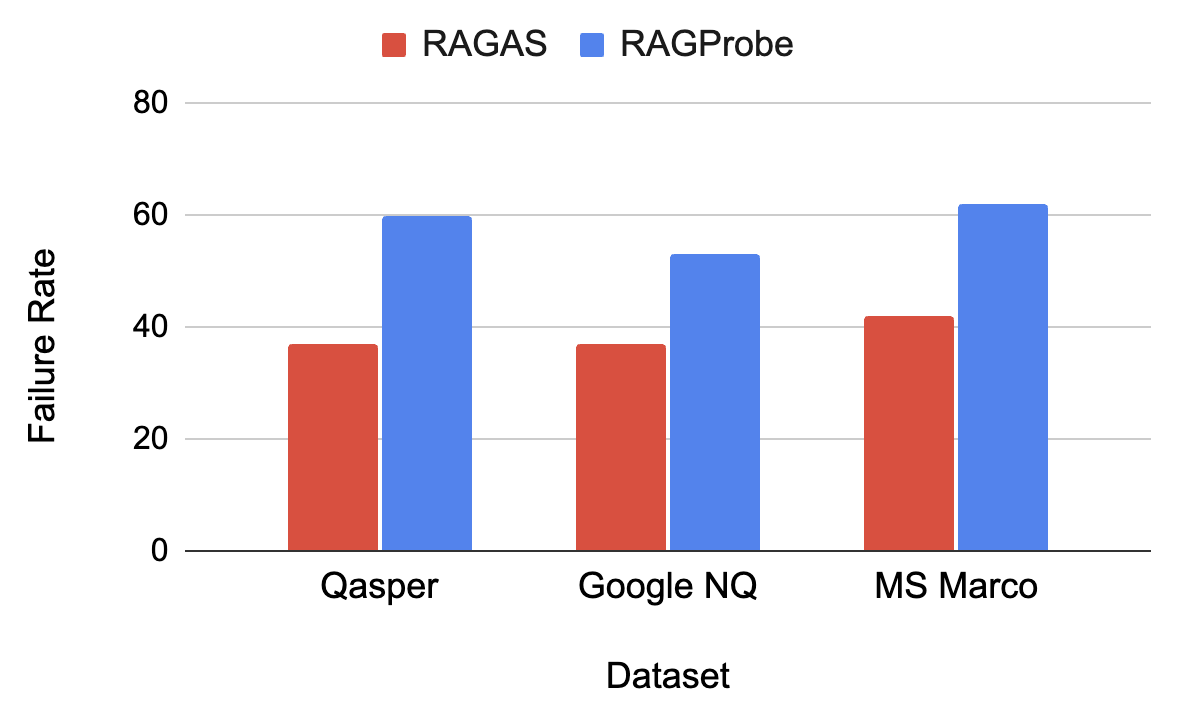}
\caption{Comparison of failure rates across datasets. A higher failure rate is better as it reveals more limitations of the RAG pipelines.} \label{fig:failure_rate_dataset}
\vspace{-3mm}
\end{figure}

\begin{figure}[htb]
\centering
\includegraphics[width=\linewidth]{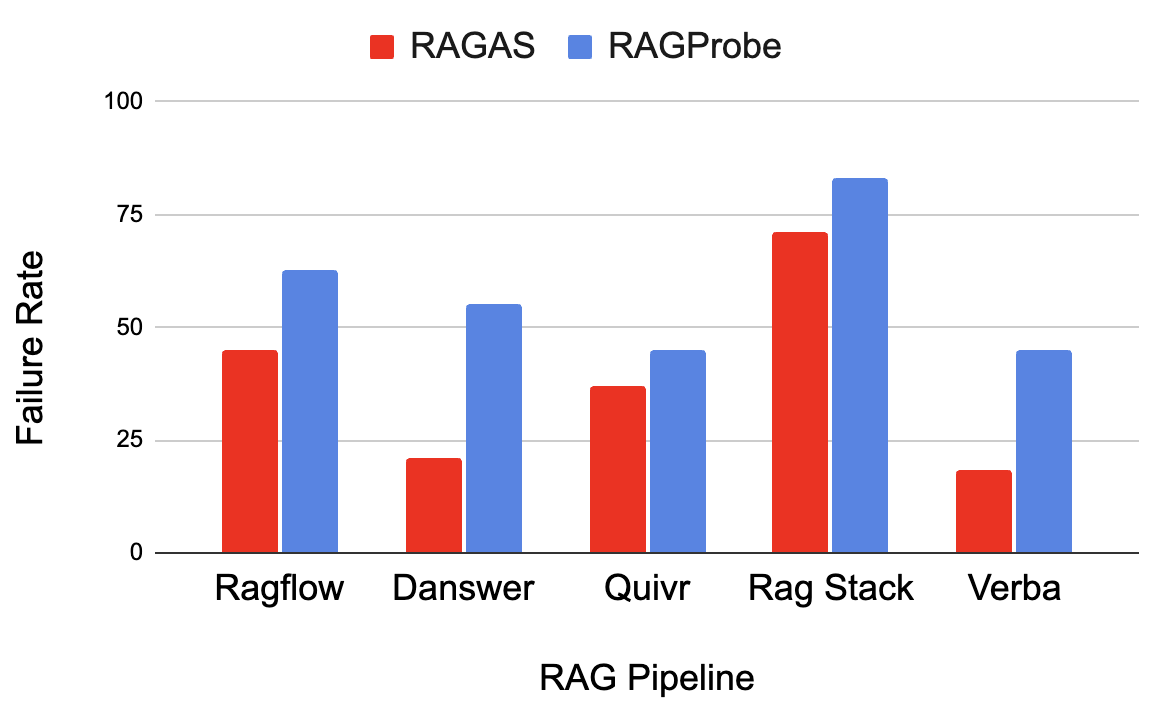}
\caption{Comparison of failure rates across RAG pipelines. A higher failure rate is better as it reveals more limitations of the RAG pipelines.}    
\label{fig:failure_rate_rag_pipeline}
% \vspace{-2mm}
\end{figure}

Our approach, compared to the state-of-the-art approach, exposed increased failure rates across each dataset and across each RAG pipeline. As shown in \autoref{fig:failure_rate_dataset}, the state-of-the-art approach exposed 37\%, 37\%, and 42\% failure rates, whereas our approach exposed 60\%, 53\%, and 62\% failure rates for Qasper, Google NQ, and MS Marco datasets respectively. The state-of-the-art approach revealed 37\%, 21\%, 45\%, 19\%, and 71\% failure rates whereas our approach revealed 45\%, 55\%, 63\%, 45\%, and 83\% failure rates across Quivr, Danswer, Ragflow, Verba, and Rag-stack respectively, as can be seen in \autoref{fig:failure_rate_rag_pipeline}. When performed the Wilcoxon signed rank test on failure rate pairs, it resulted in p-values of 0.25, and 0.0625 per dataset and per RAG pipeline respectively. Both p-values being higher than the common significance level of 0.05, suggesting that there is no significant difference in the failure rates across different datasets and across different RAG pipelines. This implies that the observed differences in failure rates might be due to random chance rather than inherent differences in the datasets or RAG pipelines being evaluated. Consequently, our evaluation shows that the performance of the RAG pipelines is relatively consistent across different datasets and pipeline implementations.

%\begin{table}[h]
%\centering
%\begin{tabular}   
%{p{.3\linewidth}|p{.3\linewidth}|p{.3\linewidth}}
%\hline
%\textbf{Dataset} & \textbf{RAGAS \cite{es2023ragas}} & \textbf{\toolname} \\
%\hline
%Qasper & 37\% & 60\% \\
%Google NQ & 37\% & 53\% \\
%MS Marco & 42\% & 62\% \\
%\hline
%\end{tabular}
%\caption{Comparison of failure rates across datasets}
%\label{tab:failure_rate_dataset}
%\vspace{-5mm}
%\end{table}

%\begin{table}[h]
%\centering
%\begin{tabular}   
%{p{.3\linewidth}|p{.3\linewidth}|p{.3\linewidth}}
%\hline
%\textbf{RAG pipeline} & \textbf{RAGAS \cite{es2023ragas}} & \textbf{\toolname} \\
%\hline
%Quivr & 37\% & 45\% \\
%Danswer & 21\% & 55\% \\
%Ragflow & 45\% & 63\% \\
%Verba & 19\% & 45\% \\
%Rag-stack & 71\% & 83\% \\
%\hline
%\end{tabular}
%\caption{Comparison of failure rates across RAG pipelines}
%\label{tab:failure_rate_rag_pipeline}
%\vspace{-5mm}
%\end{table}

\autoref{tab:validity_questions} compares the validity of questions generated by RAGAS and our approach by evaluating the similarity between pre-assigned chunks for generated questions and retrieved chunks. For the Qasper dataset, RAGAS achieved an 87\% validity rate, while our approach achieved 90\%. For the Google NQ dataset, RAGAS had a 93\% validity rate, compared to 98\% with our approach. For the MS Marco dataset, RAGAS achieved an 85\% validity rate, while our approach reached 92\%. This demonstrates that our approach consistently outperforms RAGAS in generating valid questions across all datasets.

%Out of 180 questions generated by RAGAS, 24 questions had answers listed as ``nan" (i.e., without valid answers). This shows that RAGAS have limitations in accurately generating valid answers and does not produce valid answers, which will affect the utility of the generated data for downstream tasks. 

%and InspectorRAGet \cite{fadnis2024inspectorraget}. 
%and InspectorRAGet \cite{fadnis2024inspectorraget} needed the entire document corpus specified inside a JSON file, which was not feasible. 

\begin{table}[h]
\centering
\begin{tabular}   
{|l|l|l|}
\hline
\textbf{Dataset} & \textbf{RAGAS \cite{es2023ragas}} & \textbf{\toolname} \\
\hline
Qasper & 87\% & 90\% \\
Google NQ & 93\% & 98\% \\
MS Marco & 85\% & 92\% \\
\hline
\end{tabular}
\caption{Comparison of validity of questions by comparing similarity between pre-assigned chunks for generated questions versus retrieved chunks.}
\label{tab:validity_questions}
\vspace{-2mm}
\end{table}

\begin{tcolorbox}
[width=3.3in, center upper, size=fbox, sharp corners, float]
{\textbf{Answer to RQ2:} \toolname found more failures (between 45-83\% per RAG pipeline, 180 questions) as compared to state-of-the-art. \toolname had a higher percentage (90-98\% per dataset) of valid questions than state-of-the-art.}\end{tcolorbox}

\subsubsection{What is the impact of different domains on the effectiveness of automatically generated question-answer pairs?} 

\begin{figure}[h]
    \centering
    \includegraphics[width=\linewidth]{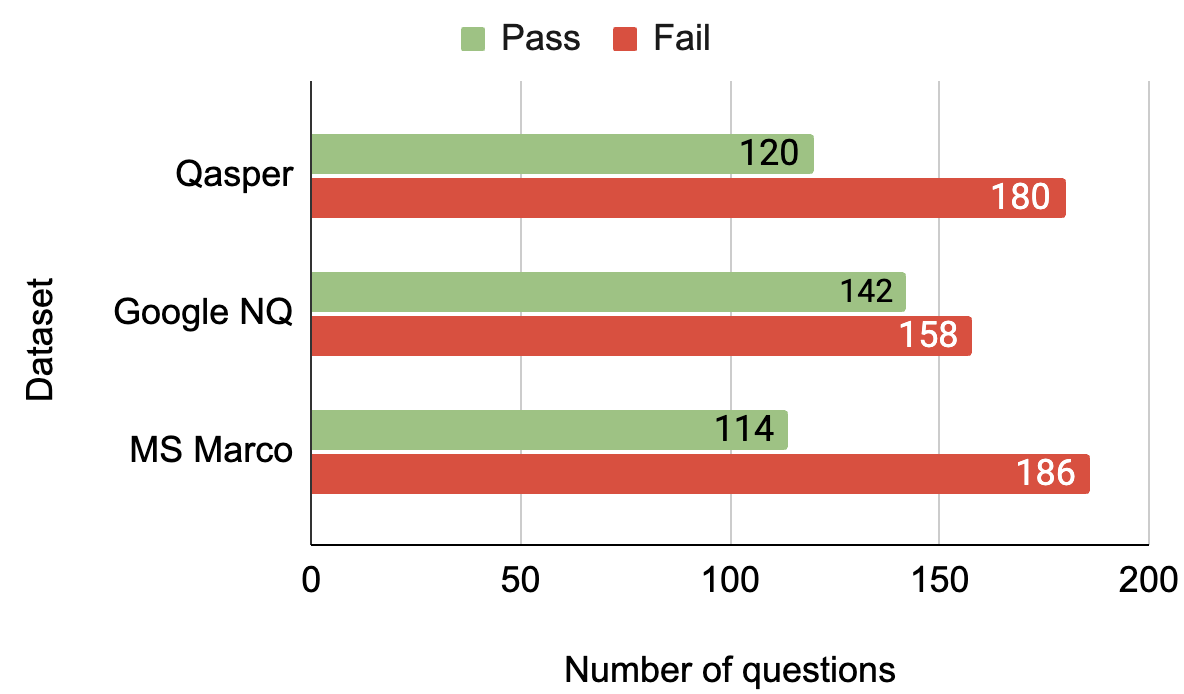}
    \caption{Automated evaluation results per dataset.}
    \label{fig:evaluation_dataset}
\end{figure}

As shown in \autoref{fig:evaluation_dataset}, MS Marco had the highest number of failing questions, whilst Google NQ had the lowest number of failing questions. 60\%, 53\%, and 62\% failure rates were observed for Qasper, Google NQ, and MS Marco datasets respectively. The high failure rates across all datasets indicate that RAG pipelines struggle consistently with the complexity and variability of questions in both academic and open-domain datasets. 

\autoref{tab:total_dataset_test_scenario} shows the number of RAG pipelines which have failed to produce correct responses across evaluation scenarios for each dataset. In summary, Google NQ and MS Marco have the highest number of failing scenarios, where all RAG pipelines failed in 5 scenarios. Qasper shows more variability, with one RAG pipeline passing in S2 and S6 scenarios.

%MS Marco notably has a lower score in S3, indicating more failures in extracting answers for that scenario.

\begin{table}[h]
\centering
\begin{tabular}   
{|c|l|l|l|}
\hline
\textbf{ID} & \textbf{Qasper} & \textbf{Google NQ} & \textbf{MS Marco} \\
\hline
S1 & 5/5 & 4/5 & 5/5 \\
S2 & 4/5 & 5/5 & 5/5 \\
S3 & 5/5 & 5/5 & 3/5 \\
S4 & 5/5 & 5/5 & 5/5 \\
S5 & 5/5 & 5/5 & 5/5  \\
S6 & 4/5 & 5/5 & 5/5 \\
\hline
\end{tabular}
\caption{Number of RAG pipelines failed across evaluation scenarios and datasets}
\label{tab:total_dataset_test_scenario}
\vspace{-5mm}
\end{table}

%As shown in \autoref{fig:evaluation_dataset}, MS Marco had the highest number of failing questions, where 121 out of 180 questions had incorrect responses. Qasper had the lowest number of failing questions, where only 88 out of 180 questions had incorrect responses. This shows that academic 

%\begin{table}[h]
%\centering
%\begin{tabular}   
%{|c|p{.25\linewidth}|p{.25\linewidth}|p{.25\linewidth}|}
%\hline
%\textbf{ID} & \textbf{Qasper} & \textbf{Google NQ} & \textbf{MS Marco} \\
%\hline
%S1 & All & Danswer, Ragflow, Verba, Rag-stack & All \\
%S2 & Danswer, Ragflow, Verba, Rag-stack & All & All \\
%S3 & All & All & Ragflow, Verba, Rag-stack \\
%S4 & All & All & All \\
%S5 & All & All & All  \\
%S6 & Danswer, Ragflow, Verba, Rag-stack & All & All \\
%\hline
%\end{tabular}
%\caption{RAG pipelines failed across evaluation scenarios and datasets. ``All" indicates that every RAG pipeline failed for a specific scenario with questions generated from the given dataset.}
%\label{tab:total_dataset_rag_pipeline_test_scenario}
%\vspace{-5mm}
%\end{table}

\begin{tcolorbox}
[width=3.3in, center upper, size=fbox, sharp corners, float]
{\textbf{Answer to RQ3:} Overall, RAG pipelines have difficulty across both academic and open-domain datasets, with a slightly better performance on open-domain questions compared to academic questions, but still exhibit high failure rates across all datasets.}\end{tcolorbox}

\section{Discussion}
%\subsection{Testing Semantic Applications}

\subsection{Evaluating RAG Pipelines}
%In this paper, we present a novel approach to generate semantic test data to evaluate semantic applications. 

The software development of RAG pipelines in practice involves making many decisions that ultimately contribute towards its quality, i.e., about text splitter, chunk size, overlap size, embedding model, large language model, vector store, distance metric for semantic similarity, top-k to rerank, reranker model, top-k to context, and prompt engineering. In real-world settings, these decisions are not grounded in methodologically sound evaluation practices but rather ad-hoc and driven by developers and product owners, constrained by project deadlines. Therefore, rigorous and robust evaluation of RAG pipelines plays an important role by having variations of questions and by automating the entire process. %differences in underlying conditions
Comparisons between RAG pipelines are unwarranted because they involve differences in their configurations, such as using different LLMs and embedding techniques. These variations mean that differences cannot be directly attributed to the pipelines themselves, but rather to the distinct underlying technologies used.

%Results from RQ1 reveal that more than 50\% of test cases fail in the following scenarios: 1) asking a question that combines multiple questions from a single document, 2) asking a question that combines multiple questions from multiple documents, and 3) asking a question for which the answer is not in the corpus. 

Results from RQ1 reveal that more than 50\% of questions fail in the following scenarios: 1) a question combining multiple questions for which answers are in a single document, 2) a question combining multiple questions for which answers are in a set of documents, and 3) a question for which answer is not in the document corpus. This highlights the necessity for thorough evaluation when deploying RAG pipelines into production. Results from RQ2 show that our approach has increased failure rates, compared to the state-of-the-art approach. Results from RQ3 show high failure rates across both academic (Qasper) and open-domain (Google NQ and MS Marco) datasets, and suggests that these pipelines struggle to accurately process and respond to a wide range of questions. Notably, while there is a marginally better performance on open-domain questions compared to academic questions, the overall high failure rates indicate that improvements are needed in the robustness of RAG pipelines. 

\subsection{Threats to Validity}
\subsubsection{Internal Validity}
%We have used LLMs to generate test questions. However, we have not validated whether the LLM-generated questions are valid. To mitigate this, we could implement a verification prompt that checks the generated questions, but we have left this for future work. Additionally, 

When evaluating the RAG generated responses against the expected responses, we considered different evaluation metrics such as correctness, relevance, completeness, consistency, explicitness, contradiction, and no-question related information. However, other evaluation metrics such as faithfulness and bias were not included. To address this, the existing Semantic Answer Evaluator (as shown in \autoref{fig:proposed_approach}) can be extended by defining additional metrics to evaluate the responses.

\subsubsection{External Validity}
In the current study, we have considered only one LLM (i.e., GPT) from OpenAI. There are other LLMs such as LLaMA, Alpaca, Vicuna, and Falcon. This limits the generalizability of our results. Investigating additional LLMs would provide a more comprehensive understanding and improve the external validity of our findings.

\subsubsection{Construct Validity}
The RAG pipelines we investigated employ various large language models, embedding models, and chunking strategies. Comparing results across these differing default settings ensures that the variability in these configurations does not bias our findings. However, having a consistent setting with the same large language model, embedding model, and chunking strategy might have yielded different results. To generate question-answer pairs, we have not considered other prompting techniques such as chain-of-thought, least-to-most, tree-of-thought, and graph-of-thought. These techniques could have produced different patterns in question-answer pairs. Exploring different prompting techniques is left for future work.

\subsection{Implications}

%We discuss the implications of our study on researchers and developers below:

\subsubsection{Researchers}
Based on identified evaluation scenarios, researchers can explore additional evaluation scenarios to create more variations in question-answer pairs. Further, researchers can include additional evaluation metrics such as faithfulness and context relevance \cite{es2023ragas, saad2023ares} for evaluating answers. Commercially available large language models such as GPT-3.5 and GPT-4 were selected for the purpose of this study based on their popularity (i.e. under extensive investigation in recent research) \cite{white2023chatgpt, wei2022chain, hou2023large, zhu2023large}. Open-source LLMs were excluded from the study as they (i) underperform \cite{zhao2023survey}, (ii) require extensive infrastructure to operate, and (iii) have been trained on smaller datasets \cite{bommasani2021opportunities}. However, researchers can investigate how open-source LLMs could perform when they are utilised to generate question-answer pairs. 

%Based on identified evaluation scenarios, researchers can investigate the impact of the scenarios which are not considered in this paper, and presented in \autoref{sec:test_scenarios}. 

%Open source LLMs such as Llama-2, Llama-3, Bloom, Falcon-7B and Vicuna-13B

%\footnote{https://openai.com/}
%Also, researchers can study the patterns on how failures are prevalent for each evaluation scenario when RAG pipelines are tested/evaluated. 

\subsubsection{Developers}
As shown in \autoref{fig:proposed_approach}, developers can consider the proposed approach to evaluate RAG pipelines that they are developing, and pinpointing the failure points. As can be seen in \autoref{tab:proposed_solutions}, developers can apply suggested fixes based on failing question-answer pairs. Automated evaluation for RAG pipelines offers benefits, including increased efficiency by reducing manual effort, ensuring consistent and reliable evaluation by minimizing human errors, and enabling integration into CI/CD pipelines. Understanding and fixing these failure points is essential for the development of effective and robust RAG pipelines. Our approach provides a guidance for developers for evaluating RAG pipelines in information retrieval and generation contexts. 

%Automated evaluation for RAG pipelines has benefits of : 1) efficiency by reducing manual effort, 2) consistent and reliable evaluating by reducing human errors, 3) integration into CI/CD pipelines to trigger test execution upon codebase changes, with automatic creation of JIRA tickets for failures (ideally, including suggested fixes within the JIRA tickets), and 4) increased test coverage because automation allows creation of numerous question-answer pairs, leading to broader test coverage. 

%3) regression testing by quickly re-executing generated questions to ensure updates to RAG pipelines do not break functionality

\begin{table}[h]
\centering
\begin{tabular}   
{|c|l|}
\hline
\textbf{ID} & \textbf{Recommendation of Fixes} \\
\hline
S1 & Upgrading embedding model \cite{gao2023retrieval, ma2023query} \\
S2 & Upgrading embedding model \cite{gao2023retrieval, ma2023query} \\
S3 & Upgrading LLM \cite{balaguer2024rag}, Fine-tuning \cite{siriwardhana2021fine, balaguer2024rag, gao2023retrieval} \\
S4 & Query rewriting \cite{ma2023query, li2024matching, tan2024small} \\
S5 & Query rewriting \cite{ma2023query, li2024matching, tan2024small} \\
S6 & Prompt engineering during generation \cite{li2024matching, gao2023retrieval} \\
\hline
\end{tabular}
\caption{Overview of evaluation scenarios with the recommended fixes based on the literature}
\label{tab:proposed_solutions}
\vspace{-5mm}
\end{table}

\section{Related Work}

%We discuss works related to evaluation and enhancement of RAG systems. %and InspectorRAGet \cite{fadnis2024inspectorraget}
Prior research proposed reference-free evaluation, such as RAGAS \cite{es2023ragas} and ARES \cite{saad2023ares}. Similar to our study, these approaches also use LLM-generated data to evaluate RAG pipelines. Previous studies \cite{es2023ragas, saad2023ares} evaluate RAG pipelines with different evaluation metrics such as contextual relevance, faithfulness, and answer relevance, whereas in our study, we assess RAG responses based on evaluation metrics such as correctness, relevance, completeness, consistency, explicit, contradiction, and no-question related information. ARES \cite{saad2023ares} utilizes a lightweight language model trained on synthetic data to assess RAG components and incorporates human-annotated data for prediction-powered inference (PPI) to mitigate errors. InspectorRAGet \cite{fadnis2024inspectorraget} is a RAG evaluation lifecycle, which considers RAG pipelines as one of the inputs, similar to our study. However, the authors \cite{fadnis2024inspectorraget} have not considered generating questions. Previous studies \cite{fadnis2024inspectorraget, chen2024benchmarking} have considered evaluation metrics which are generic to evaluating LLM responses and not specific to evaluating RAG responses. %Recent approach
Codium-AI \cite{alshahwan2024automated} generates unit tests using LLMs and requires an initial set of unit tests from developers. In contrast, our approach generates question-answer pairs from scratch using a corpus of documents. Since Codium-AI generates unit tests based on an initial set, there is a risk of propagating errors if the initial unit tests are flawed. Also, Codium-AI offers four prompt templates for generating tests, whereas our approach provides six evaluation scenarios for developers to choose from. 

%existing methods for evaluating effectiveness of RAG pipelines
The effectiveness of RAG pipelines is assessed through various methods, including evaluating individual components and their integration, using benchmarks, and applying different evaluation frameworks. For instance, PipeRAG \cite{jiang2024piperag} system integrates pipeline parallelism, flexible retrieval intervals, and a performance model to reduce generation latency and enhance quality. The eRAG method \cite{salemi2024evaluating} evaluates document relevance by utilizing each document individually and comparing the output against downstream task ground truth labels. ClapNQ \cite{rosenthal2024clapnq} provides a benchmark dataset focusing on long-form question answering to highlight areas for improvement in RAG pipelines. Comprehensive benchmarks evaluating components in RAG pipelines in various scenarios are crucial for a thorough assessment, addressing limitations like outdated information. Another research \cite{cuconasu2024power} evaluates information retrieval component of RAG pipelines and found that having irrelevant documents in the corpus can unexpectedly improve performance of a RAG pipeline. Tang et al. \cite{tang2024multihop} developed a dataset with multi-hop queries to evaluate RAG pipelines, while our approach focuses on evaluating RAG pipelines with a diverse range of scenarios.

\section{Conclusion and Future Work}

%Conclusion
In this paper, we present a novel approach for evaluating RAG pipelines through evaluation scenarios. We implemented our approach in \toolname and evaluated against 5 open-source RAG implementations. Our approach utilises six different evaluation scenarios to expose limitations in RAG pipelines. Additionally, our approach outperformed the existing state-of-the-art methods by demonstrating higher failure rates across each RAG pipeline and dataset. This addresses a critical aspect of robust evaluation by introducing challenging instances that mirror real-world complexities. 

%Future work
Future work involves integrating more evaluation scenarios, assessing flakiness by executing generated questions multiple times, and providing suggested fixes (mined from the literature and empirically validated). Further, conducting ablation studies to investigate the impact of chosen configurations (e.g. large language model, embedding model, context size, chunk size, chunking strategy) for RAG pipelines is left for our future work. We also need to generalise the RAG Evaluation Runner to be independent of the RAG implementation. The internals of a RAG pipeline are not known to the developer during evaluation. Therefore, irrespective of internal configurations, identifying evaluation scenarios and generating question-answer pairs for each scenario plays a vital role in evaluating RAG pipelines.

%The proposed automation framework only consists of generating test cases for six test scenarios. In our future work, we plan to automate generating test cases for other identified test scenarios. 

%\section{Authors and Affiliations}

%Grouping authors' names or e-mail addresses, or providing an ``e-mail
%alias,'' as shown below, is not acceptable:
%\begin{verbatim}
%  \author{Brooke Aster, David Mehldau}
%  \email{dave,judy,steve@university.edu}
%  \email{firstname.lastname@phillips.org}
%\end{verbatim}

%\section{Conclusion}

%\section{Citations and Bibliographies}

%please include this command in the {\bfseries preamble} (before the command
%``\verb|\begin{document}|'') of your \LaTeX\ source:
%\begin{verbatim}
%  \citestyle{acmauthoryear}
%\end{verbatim}

% Some examples.  A paginated journal article \cite{Abril07}, an
%  enumerated journal article \cite{Cohen07}, a reference to an entire
%  issue \cite{JCohen96}, a monograph (whole book) \cite{Kosiur01}

%\section{Acknowledgments}

%This section has a special environment:
%\begin{verbatim}
%  \begin{acks}
%  ...
%  \end{acks}
%\end{verbatim}

%\section{Appendices}

%Start the appendix with the ``\verb|appendix|'' command:
%\begin{verbatim}
%  \appendix
%\end{verbatim}

%%
%% The acknowledgments section is defined using the "acks" environment
%% (and NOT an unnumbered section). This ensures the proper
%% identification of the section in the article metadata, and the
%% consistent spelling of the heading.
%\begin{acks}
%To Robert, for the bagels and explaining CMYK and color spaces.
%\end{acks}

%%
%% The next two lines define the bibliography style to be used, and
%% the bibliography file.
\bibliographystyle{ACM-Reference-Format}
\bibliography{references}

%%% -*-BibTeX-*-
%%% Do NOT edit. File created by BibTeX with style
%%% ACM-Reference-Format-Journals [18-Jan-2012].

\begin{thebibliography}{39}

%%% ====================================================================
%%% NOTE TO THE USER: you can override these defaults by providing
%%% customized versions of any of these macros before the \bibliography
%%% command.  Each of them MUST provide its own final punctuation,
%%% except for \shownote{}, \showDOI{}, and \showURL{}.  The latter two
%%% do not use final punctuation, in order to avoid confusing it with
%%% the Web address.
%%%
%%% To suppress output of a particular field, define its macro to expand
%%% to an empty string, or better, \unskip, like this:
%%%
%%% \newcommand{\showDOI}[1]{\unskip}   % LaTeX syntax
%%%
%%% \def \showDOI #1{\unskip}           % plain TeX syntax
%%%
%%% ====================================================================

\ifx \showCODEN    \undefined \def \showCODEN     #1{\unskip}     \fi
\ifx \showDOI      \undefined \def \showDOI       #1{#1}\fi
\ifx \showISBNx    \undefined \def \showISBNx     #1{\unskip}     \fi
\ifx \showISBNxiii \undefined \def \showISBNxiii  #1{\unskip}     \fi
\ifx \showISSN     \undefined \def \showISSN      #1{\unskip}     \fi
\ifx \showLCCN     \undefined \def \showLCCN      #1{\unskip}     \fi
\ifx \shownote     \undefined \def \shownote      #1{#1}          \fi
\ifx \showarticletitle \undefined \def \showarticletitle #1{#1}   \fi
\ifx \showURL      \undefined \def \showURL       {\relax}        \fi
% The following commands are used for tagged output and should be
% invisible to TeX
\providecommand\bibfield[2]{#2}
\providecommand\bibinfo[2]{#2}
\providecommand\natexlab[1]{#1}
\providecommand\showeprint[2][]{arXiv:#2}

\bibitem[Alshahwan et~al\mbox{.}(2024)]%
        {alshahwan2024automated}
\bibfield{author}{\bibinfo{person}{Nadia Alshahwan}, \bibinfo{person}{Jubin Chheda}, \bibinfo{person}{Anastasia Finegenova}, \bibinfo{person}{Beliz Gokkaya}, \bibinfo{person}{Mark Harman}, \bibinfo{person}{Inna Harper}, \bibinfo{person}{Alexandru Marginean}, \bibinfo{person}{Shubho Sengupta}, {and} \bibinfo{person}{Eddy Wang}.} \bibinfo{year}{2024}\natexlab{}.
\newblock \showarticletitle{Automated unit test improvement using large language models at meta}.
\newblock \bibinfo{journal}{\emph{arXiv preprint arXiv:2402.09171}} (\bibinfo{year}{2024}).
\newblock


\bibitem[Asai et~al\mbox{.}(2023)]%
        {asai2023self}
\bibfield{author}{\bibinfo{person}{Akari Asai}, \bibinfo{person}{Zeqiu Wu}, \bibinfo{person}{Yizhong Wang}, \bibinfo{person}{Avirup Sil}, {and} \bibinfo{person}{Hannaneh Hajishirzi}.} \bibinfo{year}{2023}\natexlab{}.
\newblock \showarticletitle{Self-rag: Learning to retrieve, generate, and critique through self-reflection}.
\newblock \bibinfo{journal}{\emph{arXiv preprint arXiv:2310.11511}} (\bibinfo{year}{2023}).
\newblock


\bibitem[Balaguer et~al\mbox{.}(2024)]%
        {balaguer2024rag}
\bibfield{author}{\bibinfo{person}{Angels Balaguer}, \bibinfo{person}{Vinamra Benara}, \bibinfo{person}{Renato~Luiz de Freitas~Cunha}, \bibinfo{person}{Roberto de~M Estev{\~a}o~Filho}, \bibinfo{person}{Todd Hendry}, \bibinfo{person}{Daniel Holstein}, \bibinfo{person}{Jennifer Marsman}, \bibinfo{person}{Nick Mecklenburg}, \bibinfo{person}{Sara Malvar}, \bibinfo{person}{Leonardo~O Nunes}, {et~al\mbox{.}}} \bibinfo{year}{2024}\natexlab{}.
\newblock \showarticletitle{RAG vs Fine-tuning: Pipelines, Tradeoffs, and a Case Study on Agriculture}.
\newblock \bibinfo{journal}{\emph{arXiv e-prints}} (\bibinfo{year}{2024}), \bibinfo{pages}{arXiv--2401}.
\newblock


\bibitem[Barnett et~al\mbox{.}(2024)]%
        {barnett2024seven}
\bibfield{author}{\bibinfo{person}{Scott Barnett}, \bibinfo{person}{Stefanus Kurniawan}, \bibinfo{person}{Srikanth Thudumu}, \bibinfo{person}{Zach Brannelly}, {and} \bibinfo{person}{Mohamed Abdelrazek}.} \bibinfo{year}{2024}\natexlab{}.
\newblock \showarticletitle{Seven Failure Points When Engineering a Retrieval Augmented Generation System}.
\newblock \bibinfo{journal}{\emph{arXiv preprint arXiv:2401.05856}} (\bibinfo{year}{2024}).
\newblock


\bibitem[Bommasani et~al\mbox{.}(2021)]%
        {bommasani2021opportunities}
\bibfield{author}{\bibinfo{person}{Rishi Bommasani}, \bibinfo{person}{Drew~A Hudson}, \bibinfo{person}{Ehsan Adeli}, \bibinfo{person}{Russ Altman}, \bibinfo{person}{Simran Arora}, \bibinfo{person}{Sydney von Arx}, \bibinfo{person}{Michael~S Bernstein}, \bibinfo{person}{Jeannette Bohg}, \bibinfo{person}{Antoine Bosselut}, \bibinfo{person}{Emma Brunskill}, {et~al\mbox{.}}} \bibinfo{year}{2021}\natexlab{}.
\newblock \showarticletitle{On the opportunities and risks of foundation models}.
\newblock \bibinfo{journal}{\emph{arXiv preprint arXiv:2108.07258}} (\bibinfo{year}{2021}).
\newblock


\bibitem[Chen et~al\mbox{.}(2024b)]%
        {chen2024benchmarking}
\bibfield{author}{\bibinfo{person}{Jiawei Chen}, \bibinfo{person}{Hongyu Lin}, \bibinfo{person}{Xianpei Han}, {and} \bibinfo{person}{Le Sun}.} \bibinfo{year}{2024}\natexlab{b}.
\newblock \showarticletitle{Benchmarking large language models in retrieval-augmented generation}. In \bibinfo{booktitle}{\emph{Proceedings of the AAAI Conference on Artificial Intelligence}}, Vol.~\bibinfo{volume}{38}. \bibinfo{pages}{17754--17762}.
\newblock


\bibitem[Chen et~al\mbox{.}(2024a)]%
        {chen2024few}
\bibfield{author}{\bibinfo{person}{Mingda Chen}, \bibinfo{person}{Xilun Chen}, {and} \bibinfo{person}{Wen-tau Yih}.} \bibinfo{year}{2024}\natexlab{a}.
\newblock \showarticletitle{Few-Shot Data Synthesis for Open Domain Multi-Hop Question Answering}. In \bibinfo{booktitle}{\emph{Proceedings of the 18th Conference of the European Chapter of the Association for Computational Linguistics (Volume 1: Long Papers)}}. \bibinfo{pages}{190--208}.
\newblock


\bibitem[Cuconasu et~al\mbox{.}(2024)]%
        {cuconasu2024power}
\bibfield{author}{\bibinfo{person}{Florin Cuconasu}, \bibinfo{person}{Giovanni Trappolini}, \bibinfo{person}{Federico Siciliano}, \bibinfo{person}{Simone Filice}, \bibinfo{person}{Cesare Campagnano}, \bibinfo{person}{Yoelle Maarek}, \bibinfo{person}{Nicola Tonellotto}, {and} \bibinfo{person}{Fabrizio Silvestri}.} \bibinfo{year}{2024}\natexlab{}.
\newblock \showarticletitle{The Power of Noise: Redefining Retrieval for RAG Systems}.
\newblock \bibinfo{journal}{\emph{arXiv preprint arXiv:2401.14887}} (\bibinfo{year}{2024}).
\newblock


\bibitem[Dasigi et~al\mbox{.}(2021)]%
        {dasigi2021dataset}
\bibfield{author}{\bibinfo{person}{Pradeep Dasigi}, \bibinfo{person}{Kyle Lo}, \bibinfo{person}{Iz Beltagy}, \bibinfo{person}{Arman Cohan}, \bibinfo{person}{Noah~A Smith}, {and} \bibinfo{person}{Matt Gardner}.} \bibinfo{year}{2021}\natexlab{}.
\newblock \showarticletitle{A dataset of information-seeking questions and answers anchored in research papers}.
\newblock \bibinfo{journal}{\emph{arXiv preprint arXiv:2105.03011}} (\bibinfo{year}{2021}).
\newblock


\bibitem[Es et~al\mbox{.}(2023)]%
        {es2023ragas}
\bibfield{author}{\bibinfo{person}{Shahul Es}, \bibinfo{person}{Jithin James}, \bibinfo{person}{Luis Espinosa-Anke}, {and} \bibinfo{person}{Steven Schockaert}.} \bibinfo{year}{2023}\natexlab{}.
\newblock \showarticletitle{Ragas: Automated evaluation of retrieval augmented generation}.
\newblock \bibinfo{journal}{\emph{arXiv preprint arXiv:2309.15217}} (\bibinfo{year}{2023}).
\newblock


\bibitem[Fadnis et~al\mbox{.}(2024)]%
        {fadnis2024inspectorraget}
\bibfield{author}{\bibinfo{person}{Kshitij Fadnis}, \bibinfo{person}{Siva~Sankalp Patel}, \bibinfo{person}{Odellia Boni}, \bibinfo{person}{Yannis Katsis}, \bibinfo{person}{Sara Rosenthal}, \bibinfo{person}{Benjamin Sznajder}, {and} \bibinfo{person}{Marina Danilevsky}.} \bibinfo{year}{2024}\natexlab{}.
\newblock \showarticletitle{InspectorRAGet: An Introspection Platform for RAG Evaluation}.
\newblock \bibinfo{journal}{\emph{arXiv preprint arXiv:2404.17347}} (\bibinfo{year}{2024}).
\newblock


\bibitem[Feng et~al\mbox{.}(2024)]%
        {feng2024retrieval}
\bibfield{author}{\bibinfo{person}{Zhangyin Feng}, \bibinfo{person}{Xiaocheng Feng}, \bibinfo{person}{Dezhi Zhao}, \bibinfo{person}{Maojin Yang}, {and} \bibinfo{person}{Bing Qin}.} \bibinfo{year}{2024}\natexlab{}.
\newblock \showarticletitle{Retrieval-generation synergy augmented large language models}. In \bibinfo{booktitle}{\emph{ICASSP 2024-2024 IEEE International Conference on Acoustics, Speech and Signal Processing (ICASSP)}}. IEEE, \bibinfo{pages}{11661--11665}.
\newblock


\bibitem[Gao et~al\mbox{.}(2023)]%
        {gao2023retrieval}
\bibfield{author}{\bibinfo{person}{Yunfan Gao}, \bibinfo{person}{Yun Xiong}, \bibinfo{person}{Xinyu Gao}, \bibinfo{person}{Kangxiang Jia}, \bibinfo{person}{Jinliu Pan}, \bibinfo{person}{Yuxi Bi}, \bibinfo{person}{Yi Dai}, \bibinfo{person}{Jiawei Sun}, {and} \bibinfo{person}{Haofen Wang}.} \bibinfo{year}{2023}\natexlab{}.
\newblock \showarticletitle{Retrieval-augmented generation for large language models: A survey}.
\newblock \bibinfo{journal}{\emph{arXiv preprint arXiv:2312.10997}} (\bibinfo{year}{2023}).
\newblock


\bibitem[Hou et~al\mbox{.}(2023)]%
        {hou2023large}
\bibfield{author}{\bibinfo{person}{Xinyi Hou}, \bibinfo{person}{Yanjie Zhao}, \bibinfo{person}{Yue Liu}, \bibinfo{person}{Zhou Yang}, \bibinfo{person}{Kailong Wang}, \bibinfo{person}{Li Li}, \bibinfo{person}{Xiapu Luo}, \bibinfo{person}{David Lo}, \bibinfo{person}{John Grundy}, {and} \bibinfo{person}{Haoyu Wang}.} \bibinfo{year}{2023}\natexlab{}.
\newblock \showarticletitle{Large language models for software engineering: A systematic literature review}.
\newblock \bibinfo{journal}{\emph{arXiv preprint arXiv:2308.10620}} (\bibinfo{year}{2023}).
\newblock


\bibitem[Jeong et~al\mbox{.}(2024)]%
        {jeong2024adaptive}
\bibfield{author}{\bibinfo{person}{Soyeong Jeong}, \bibinfo{person}{Jinheon Baek}, \bibinfo{person}{Sukmin Cho}, \bibinfo{person}{Sung~Ju Hwang}, {and} \bibinfo{person}{Jong~C Park}.} \bibinfo{year}{2024}\natexlab{}.
\newblock \showarticletitle{Adaptive-rag: Learning to adapt retrieval-augmented large language models through question complexity}.
\newblock \bibinfo{journal}{\emph{arXiv preprint arXiv:2403.14403}} (\bibinfo{year}{2024}).
\newblock


\bibitem[Jiang et~al\mbox{.}(2024)]%
        {jiang2024piperag}
\bibfield{author}{\bibinfo{person}{Wenqi Jiang}, \bibinfo{person}{Shuai Zhang}, \bibinfo{person}{Boran Han}, \bibinfo{person}{Jie Wang}, \bibinfo{person}{Bernie Wang}, {and} \bibinfo{person}{Tim Kraska}.} \bibinfo{year}{2024}\natexlab{}.
\newblock \showarticletitle{Piperag: Fast retrieval-augmented generation via algorithm-system co-design}.
\newblock \bibinfo{journal}{\emph{arXiv preprint arXiv:2403.05676}} (\bibinfo{year}{2024}).
\newblock


\bibitem[Kwiatkowski et~al\mbox{.}(2019)]%
        {kwiatkowski2019natural}
\bibfield{author}{\bibinfo{person}{Tom Kwiatkowski}, \bibinfo{person}{Jennimaria Palomaki}, \bibinfo{person}{Olivia Redfield}, \bibinfo{person}{Michael Collins}, \bibinfo{person}{Ankur Parikh}, \bibinfo{person}{Chris Alberti}, \bibinfo{person}{Danielle Epstein}, \bibinfo{person}{Illia Polosukhin}, \bibinfo{person}{Jacob Devlin}, \bibinfo{person}{Kenton Lee}, {et~al\mbox{.}}} \bibinfo{year}{2019}\natexlab{}.
\newblock \showarticletitle{Natural questions: a benchmark for question answering research}.
\newblock \bibinfo{journal}{\emph{Transactions of the Association for Computational Linguistics}}  \bibinfo{volume}{7} (\bibinfo{year}{2019}), \bibinfo{pages}{453--466}.
\newblock


\bibitem[Li et~al\mbox{.}(2024)]%
        {li2024matching}
\bibfield{author}{\bibinfo{person}{Xiaoxi Li}, \bibinfo{person}{Jiajie Jin}, \bibinfo{person}{Yujia Zhou}, \bibinfo{person}{Yuyao Zhang}, \bibinfo{person}{Peitian Zhang}, \bibinfo{person}{Yutao Zhu}, {and} \bibinfo{person}{Zhicheng Dou}.} \bibinfo{year}{2024}\natexlab{}.
\newblock \showarticletitle{From Matching to Generation: A Survey on Generative Information Retrieval}.
\newblock \bibinfo{journal}{\emph{arXiv preprint arXiv:2404.14851}} (\bibinfo{year}{2024}).
\newblock


\bibitem[Liu et~al\mbox{.}(2023)]%
        {liu2023gpteval}
\bibfield{author}{\bibinfo{person}{Yang Liu}, \bibinfo{person}{Dan Iter}, \bibinfo{person}{Yichong Xu}, \bibinfo{person}{Shuohang Wang}, \bibinfo{person}{Ruochen Xu}, {and} \bibinfo{person}{Chenguang Zhu}.} \bibinfo{year}{2023}\natexlab{}.
\newblock \showarticletitle{Gpteval: Nlg evaluation using gpt-4 with better human alignment}.
\newblock \bibinfo{journal}{\emph{arXiv preprint arXiv:2303.16634}} (\bibinfo{year}{2023}).
\newblock


\bibitem[Ma et~al\mbox{.}(2023)]%
        {ma2023query}
\bibfield{author}{\bibinfo{person}{Xinbei Ma}, \bibinfo{person}{Yeyun Gong}, \bibinfo{person}{Pengcheng He}, \bibinfo{person}{Hai Zhao}, {and} \bibinfo{person}{Nan Duan}.} \bibinfo{year}{2023}\natexlab{}.
\newblock \showarticletitle{Query rewriting for retrieval-augmented large language models}.
\newblock \bibinfo{journal}{\emph{arXiv preprint arXiv:2305.14283}} (\bibinfo{year}{2023}).
\newblock


\bibitem[Mao et~al\mbox{.}(2020)]%
        {mao2020generation}
\bibfield{author}{\bibinfo{person}{Yuning Mao}, \bibinfo{person}{Pengcheng He}, \bibinfo{person}{Xiaodong Liu}, \bibinfo{person}{Yelong Shen}, \bibinfo{person}{Jianfeng Gao}, \bibinfo{person}{Jiawei Han}, {and} \bibinfo{person}{Weizhu Chen}.} \bibinfo{year}{2020}\natexlab{}.
\newblock \showarticletitle{Generation-augmented retrieval for open-domain question answering}.
\newblock \bibinfo{journal}{\emph{arXiv preprint arXiv:2009.08553}} (\bibinfo{year}{2020}).
\newblock


\bibitem[Nguyen et~al\mbox{.}(2016)]%
        {nguyen2016ms}
\bibfield{author}{\bibinfo{person}{Tri Nguyen}, \bibinfo{person}{Mir Rosenberg}, \bibinfo{person}{Xia Song}, \bibinfo{person}{Jianfeng Gao}, \bibinfo{person}{Saurabh Tiwary}, \bibinfo{person}{Rangan Majumder}, {and} \bibinfo{person}{Li Deng}.} \bibinfo{year}{2016}\natexlab{}.
\newblock \showarticletitle{Ms marco: A human-generated machine reading comprehension dataset}.
\newblock  (\bibinfo{year}{2016}).
\newblock


\bibitem[OpenAI(2023)]%
        {openai2023gpt4}
\bibfield{author}{\bibinfo{person}{OpenAI}.} \bibinfo{year}{2023}\natexlab{}.
\newblock \bibinfo{title}{GPT-4 Technical Report}.
\newblock
\newblock
\showeprint[arxiv]{2303.08774}~[cs.CL]


\bibitem[Rasool et~al\mbox{.}(2023)]%
        {rasool2023evaluating}
\bibfield{author}{\bibinfo{person}{Zafaryab Rasool}, \bibinfo{person}{Scott Barnett}, \bibinfo{person}{Stefanus Kurniawan}, \bibinfo{person}{Sherwin Balugo}, \bibinfo{person}{Rajesh Vasa}, \bibinfo{person}{Courtney Chesser}, {and} \bibinfo{person}{Alex Bahar-Fuchs}.} \bibinfo{year}{2023}\natexlab{}.
\newblock \showarticletitle{Evaluating LLMs on Document-Based QA: Exact Answer Selection and Numerical Extraction using Cogtale dataset}.
\newblock \bibinfo{journal}{\emph{arXiv preprint arXiv:2311.07878}} (\bibinfo{year}{2023}).
\newblock


\bibitem[Rasool et~al\mbox{.}(2024)]%
        {rasool2024llms}
\bibfield{author}{\bibinfo{person}{Zafaryab Rasool}, \bibinfo{person}{Scott Barnett}, \bibinfo{person}{David Willie}, \bibinfo{person}{Stefanus Kurniawan}, \bibinfo{person}{Sherwin Balugo}, \bibinfo{person}{Srikanth Thudumu}, {and} \bibinfo{person}{Mohamed Abdelrazek}.} \bibinfo{year}{2024}\natexlab{}.
\newblock \showarticletitle{LLMs for Test Input Generation for Semantic Caches}.
\newblock \bibinfo{journal}{\emph{arXiv preprint arXiv:2401.08138}} (\bibinfo{year}{2024}).
\newblock


\bibitem[Rosenthal et~al\mbox{.}(2024)]%
        {rosenthal2024clapnq}
\bibfield{author}{\bibinfo{person}{Sara Rosenthal}, \bibinfo{person}{Avirup Sil}, \bibinfo{person}{Radu Florian}, {and} \bibinfo{person}{Salim Roukos}.} \bibinfo{year}{2024}\natexlab{}.
\newblock \showarticletitle{CLAPNQ: Cohesive Long-form Answers from Passages in Natural Questions for RAG systems}.
\newblock \bibinfo{journal}{\emph{arXiv preprint arXiv:2404.02103}} (\bibinfo{year}{2024}).
\newblock


\bibitem[Saad-Falcon et~al\mbox{.}(2023)]%
        {saad2023ares}
\bibfield{author}{\bibinfo{person}{Jon Saad-Falcon}, \bibinfo{person}{Omar Khattab}, \bibinfo{person}{Christopher Potts}, {and} \bibinfo{person}{Matei Zaharia}.} \bibinfo{year}{2023}\natexlab{}.
\newblock \showarticletitle{Ares: An automated evaluation framework for retrieval-augmented generation systems}.
\newblock \bibinfo{journal}{\emph{arXiv preprint arXiv:2311.09476}} (\bibinfo{year}{2023}).
\newblock


\bibitem[Salemi and Zamani(2024)]%
        {salemi2024evaluating}
\bibfield{author}{\bibinfo{person}{Alireza Salemi} {and} \bibinfo{person}{Hamed Zamani}.} \bibinfo{year}{2024}\natexlab{}.
\newblock \showarticletitle{Evaluating Retrieval Quality in Retrieval-Augmented Generation}.
\newblock \bibinfo{journal}{\emph{arXiv preprint arXiv:2404.13781}} (\bibinfo{year}{2024}).
\newblock


\bibitem[Siriwardhana et~al\mbox{.}(2021)]%
        {siriwardhana2021fine}
\bibfield{author}{\bibinfo{person}{Shamane Siriwardhana}, \bibinfo{person}{Rivindu Weerasekera}, \bibinfo{person}{Elliott Wen}, {and} \bibinfo{person}{Suranga Nanayakkara}.} \bibinfo{year}{2021}\natexlab{}.
\newblock \showarticletitle{Fine-tune the Entire RAG Architecture (including DPR retriever) for Question-Answering}.
\newblock \bibinfo{journal}{\emph{arXiv preprint arXiv:2106.11517}} (\bibinfo{year}{2021}).
\newblock


\bibitem[Tan et~al\mbox{.}(2024)]%
        {tan2024small}
\bibfield{author}{\bibinfo{person}{Jiejun Tan}, \bibinfo{person}{Zhicheng Dou}, \bibinfo{person}{Yutao Zhu}, \bibinfo{person}{Peidong Guo}, \bibinfo{person}{Kun Fang}, {and} \bibinfo{person}{Ji-Rong Wen}.} \bibinfo{year}{2024}\natexlab{}.
\newblock \showarticletitle{Small Models, Big Insights: Leveraging Slim Proxy Models To Decide When and What to Retrieve for LLMs}.
\newblock \bibinfo{journal}{\emph{arXiv preprint arXiv:2402.12052}} (\bibinfo{year}{2024}).
\newblock


\bibitem[Tang and Yang(2024)]%
        {tang2024multihop}
\bibfield{author}{\bibinfo{person}{Yixuan Tang} {and} \bibinfo{person}{Yi Yang}.} \bibinfo{year}{2024}\natexlab{}.
\newblock \showarticletitle{MultiHop-RAG: Benchmarking Retrieval-Augmented Generation for Multi-Hop Queries}.
\newblock \bibinfo{journal}{\emph{arXiv preprint arXiv:2401.15391}} (\bibinfo{year}{2024}).
\newblock


\bibitem[Wei et~al\mbox{.}(2022)]%
        {wei2022chain}
\bibfield{author}{\bibinfo{person}{Jason Wei}, \bibinfo{person}{Xuezhi Wang}, \bibinfo{person}{Dale Schuurmans}, \bibinfo{person}{Maarten Bosma}, \bibinfo{person}{Fei Xia}, \bibinfo{person}{Ed Chi}, \bibinfo{person}{Quoc~V Le}, \bibinfo{person}{Denny Zhou}, {et~al\mbox{.}}} \bibinfo{year}{2022}\natexlab{}.
\newblock \showarticletitle{Chain-of-thought prompting elicits reasoning in large language models}.
\newblock \bibinfo{journal}{\emph{Advances in Neural Information Processing Systems}}  \bibinfo{volume}{35} (\bibinfo{year}{2022}), \bibinfo{pages}{24824--24837}.
\newblock


\bibitem[White et~al\mbox{.}(2023)]%
        {white2023chatgpt}
\bibfield{author}{\bibinfo{person}{Jules White}, \bibinfo{person}{Sam Hays}, \bibinfo{person}{Quchen Fu}, \bibinfo{person}{Jesse Spencer-Smith}, {and} \bibinfo{person}{Douglas~C Schmidt}.} \bibinfo{year}{2023}\natexlab{}.
\newblock \showarticletitle{Chatgpt prompt patterns for improving code quality, refactoring, requirements elicitation, and software design}.
\newblock \bibinfo{journal}{\emph{arXiv preprint arXiv:2303.07839}} (\bibinfo{year}{2023}).
\newblock


\bibitem[Woolson(2005)]%
        {woolson2005wilcoxon}
\bibfield{author}{\bibinfo{person}{Robert~F Woolson}.} \bibinfo{year}{2005}\natexlab{}.
\newblock \showarticletitle{Wilcoxon signed-rank test}.
\newblock \bibinfo{journal}{\emph{Encyclopedia of Biostatistics}}  \bibinfo{volume}{8} (\bibinfo{year}{2005}).
\newblock


\bibitem[Wu et~al\mbox{.}(2024)]%
        {wu2024faithful}
\bibfield{author}{\bibinfo{person}{Kevin Wu}, \bibinfo{person}{Eric Wu}, {and} \bibinfo{person}{James Zou}.} \bibinfo{year}{2024}\natexlab{}.
\newblock \showarticletitle{How faithful are RAG models? Quantifying the tug-of-war between RAG and LLMs' internal prior}.
\newblock \bibinfo{journal}{\emph{arXiv preprint arXiv:2404.10198}} (\bibinfo{year}{2024}).
\newblock


\bibitem[Yan et~al\mbox{.}(2024)]%
        {yan2024corrective}
\bibfield{author}{\bibinfo{person}{Shi-Qi Yan}, \bibinfo{person}{Jia-Chen Gu}, \bibinfo{person}{Yun Zhu}, {and} \bibinfo{person}{Zhen-Hua Ling}.} \bibinfo{year}{2024}\natexlab{}.
\newblock \showarticletitle{Corrective Retrieval Augmented Generation}.
\newblock \bibinfo{journal}{\emph{arXiv preprint arXiv:2401.15884}} (\bibinfo{year}{2024}).
\newblock


\bibitem[Zhang et~al\mbox{.}(2024)]%
        {zhang2024raft}
\bibfield{author}{\bibinfo{person}{Tianjun Zhang}, \bibinfo{person}{Shishir~G Patil}, \bibinfo{person}{Naman Jain}, \bibinfo{person}{Sheng Shen}, \bibinfo{person}{Matei Zaharia}, \bibinfo{person}{Ion Stoica}, {and} \bibinfo{person}{Joseph~E Gonzalez}.} \bibinfo{year}{2024}\natexlab{}.
\newblock \showarticletitle{Raft: Adapting language model to domain specific rag}.
\newblock \bibinfo{journal}{\emph{arXiv preprint arXiv:2403.10131}} (\bibinfo{year}{2024}).
\newblock


\bibitem[Zhao et~al\mbox{.}(2023)]%
        {zhao2023survey}
\bibfield{author}{\bibinfo{person}{Wayne~Xin Zhao}, \bibinfo{person}{Kun Zhou}, \bibinfo{person}{Junyi Li}, \bibinfo{person}{Tianyi Tang}, \bibinfo{person}{Xiaolei Wang}, \bibinfo{person}{Yupeng Hou}, \bibinfo{person}{Yingqian Min}, \bibinfo{person}{Beichen Zhang}, \bibinfo{person}{Junjie Zhang}, \bibinfo{person}{Zican Dong}, {et~al\mbox{.}}} \bibinfo{year}{2023}\natexlab{}.
\newblock \showarticletitle{A survey of large language models}.
\newblock \bibinfo{journal}{\emph{arXiv preprint arXiv:2303.18223}} (\bibinfo{year}{2023}).
\newblock


\bibitem[Zhu et~al\mbox{.}(2023)]%
        {zhu2023large}
\bibfield{author}{\bibinfo{person}{Zhaocheng Zhu}, \bibinfo{person}{Yuan Xue}, \bibinfo{person}{Xinyun Chen}, \bibinfo{person}{Denny Zhou}, \bibinfo{person}{Jian Tang}, \bibinfo{person}{Dale Schuurmans}, {and} \bibinfo{person}{Hanjun Dai}.} \bibinfo{year}{2023}\natexlab{}.
\newblock \showarticletitle{Large language models can learn rules}.
\newblock \bibinfo{journal}{\emph{arXiv preprint arXiv:2310.07064}} (\bibinfo{year}{2023}).
\newblock


\end{thebibliography}

%%
%% If your work has an appendix, this is the place to put it.
%\appendix

%\section{Research Methods}

%\subsection{Part One}

%\subsection{Part Two}

%\section{Online Resources}

\end{document}